\documentclass{article}

\usepackage[final,nonatbib]{neurips_2019}

\usepackage[utf8]{inputenc} % allow utf-8 input
\usepackage[T1]{fontenc}    % use 8-bit T1 fonts
\usepackage{hyperref}       % hyperlinks
\usepackage{url}            % simple URL typesetting
\usepackage{booktabs}       % professional-quality tables
\usepackage{amsfonts}       % blackboard math symbols
\usepackage{nicefrac}       % compact symbols for 1/2, etc.
\usepackage{microtype}      % microtypography

\usepackage{bm}
\usepackage{multirow}
\usepackage{amsmath}
\usepackage{amssymb}
\usepackage{amsthm}
\usepackage{color}
\usepackage{times}
\usepackage{graphicx} % more modern
\usepackage{subfigure} 
\usepackage[rightcaption]{sidecap}
\usepackage{threeparttable}
\usepackage{algorithm}
\usepackage{algorithmic}
\newtheorem{theorem}{Theorem}[section]

\newtheorem{proposition}[theorem]{Proposition}

\newtheorem{definition}[theorem]{Definition}

\title{Scalable Gromov-Wasserstein Learning for\\Graph Partitioning and Matching}

\author{
  Hongteng Xu$^{1,2}$\quad\quad Dixin Luo$^{2}$\quad\quad Lawrence Carin$^{2}$ \\
  $^1$Infinia ML Inc.\quad\quad $^2$Duke University\\
  \texttt{\{hongteng.xu, dixin.luo, lcarin\}@duke.edu} \\
}

\begin{document}

\maketitle

\begin{abstract}
We propose a scalable Gromov-Wasserstein learning (S-GWL) method and establish a novel and theoretically-supported paradigm for large-scale graph analysis.
The proposed method is based on the fact that Gromov-Wasserstein discrepancy is a pseudometric on graphs. 
Given two graphs, the optimal transport associated with their Gromov-Wasserstein discrepancy provides the correspondence between their nodes and achieves graph matching. 
When one of the graphs has isolated but self-connected nodes ($i.e.$, a disconnected graph), the optimal transport indicates the clustering structure of the other graph and achieves graph partitioning. 
Using this concept, we extend our method to multi-graph partitioning and matching by learning a Gromov-Wasserstein barycenter graph for multiple observed graphs; the barycenter graph plays the role of the disconnected graph, and since it is learned, so is the clustering.  
Our method combines a recursive $K$-partition mechanism with a regularized proximal gradient algorithm, whose time complexity is $\mathcal{O}(K(E+V)\log_K V)$ for graphs with $V$ nodes and $E$ edges. 
To our knowledge, our method is the first attempt to make Gromov-Wasserstein discrepancy applicable to large-scale graph analysis and unify graph partitioning and matching into the same framework.
It outperforms state-of-the-art graph partitioning and matching methods, achieving a trade-off between accuracy and efficiency. 
\end{abstract}

\section{Introduction}
\vspace{-8pt}
Gromov-Wasserstein distance~\cite{sturm2006geometry,memoli2011gromov} was originally designed for metric-measure spaces, which can measure distances between distributions in a relational way, deriving an optimal transport between the samples in distinct spaces. 
Recently, the work in~\cite{chowdhury2018gromov} proved that this distance can be extended to \emph{Gromov-Wasserstein discrepancy} (GW discrepancy)~\cite{peyre2016gromov}, which defines a pseudometric for graphs. 
Accordingly, the optimal transport between two graphs indicates the correspondence between their nodes. 
This work theoretically supports the applications of GW discrepancy to structural data analysis, $e.g.$, 2D/3D object matching~\cite{memoli2004comparing,memoli2009spectral,bronstein2010gromov}, molecule analysis~\cite{vayer2018fused,vayer2018optimal}, network alignment~\cite{xu2019gromov}, etc. 
Unfortunately, although GW discrepancy-based methods are attractive theoretically, they are often inapplicable to large-scale graphs, because of high computational complexity. 
Additionally, these methods are designed for two-graph matching, ignoring the potential of GW discrepancy to other tasks, like graph partitioning and multi-graph matching. 
As a result, the partitioning and the matching of large-scale graphs still typically rely on heuristic methods~\cite{girvan2002community,clauset2004finding,vijayan2015magna++,malod2015graal}, whose performance is often sub-optimal, especially in noisy cases. 

\begin{figure}[t]
    \centering
    \begin{minipage}[b]{0.22\linewidth}
    \centering
    \subfigure[\tiny{Graph matching}]{
    \includegraphics[height=1.5cm]{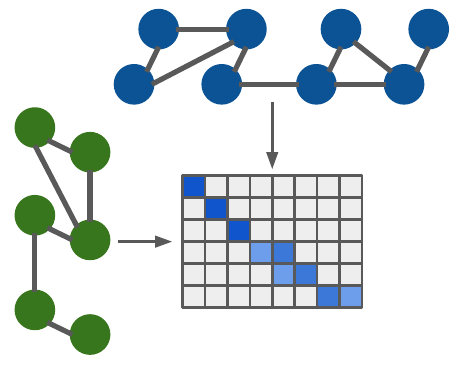}\label{fig:gm}
    }\\
    \vspace{-7pt}
    \subfigure[\tiny{Graph partitioning}]{
    \includegraphics[height=1.5cm]{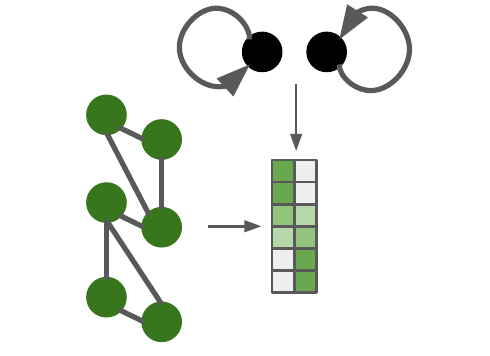}\label{fig:gp}

    }
    \end{minipage}
    \begin{minipage}[b]{0.35\linewidth}
    \centering
    \subfigure[\tiny{Multi-graph matching}]{
    \includegraphics[height=1.5cm]{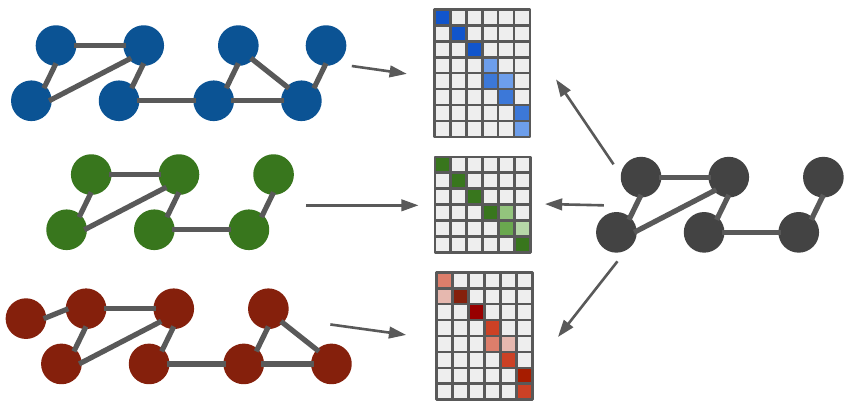}\label{fig:gms}
    }\\
    \vspace{-7pt}
    \subfigure[\tiny{Multi-graph partitioning}]{
    \includegraphics[height=1.5cm]{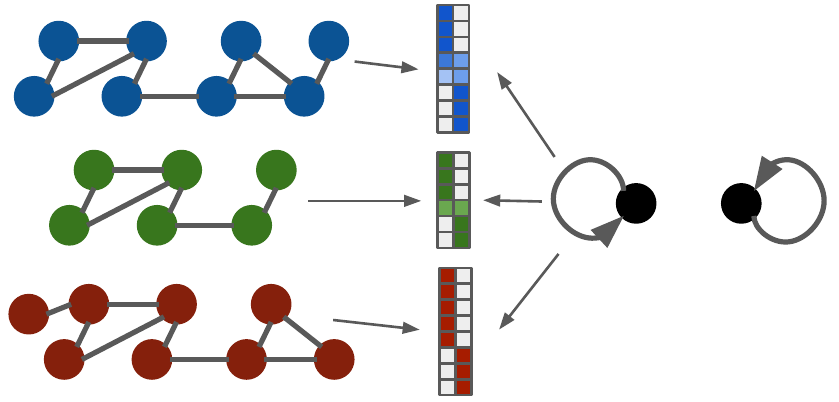}\label{fig:gps}
    }
    \end{minipage}
    \subfigure[\tiny{Comparisons on accuracy and efficiency}]{
    \includegraphics[height=3.5cm]{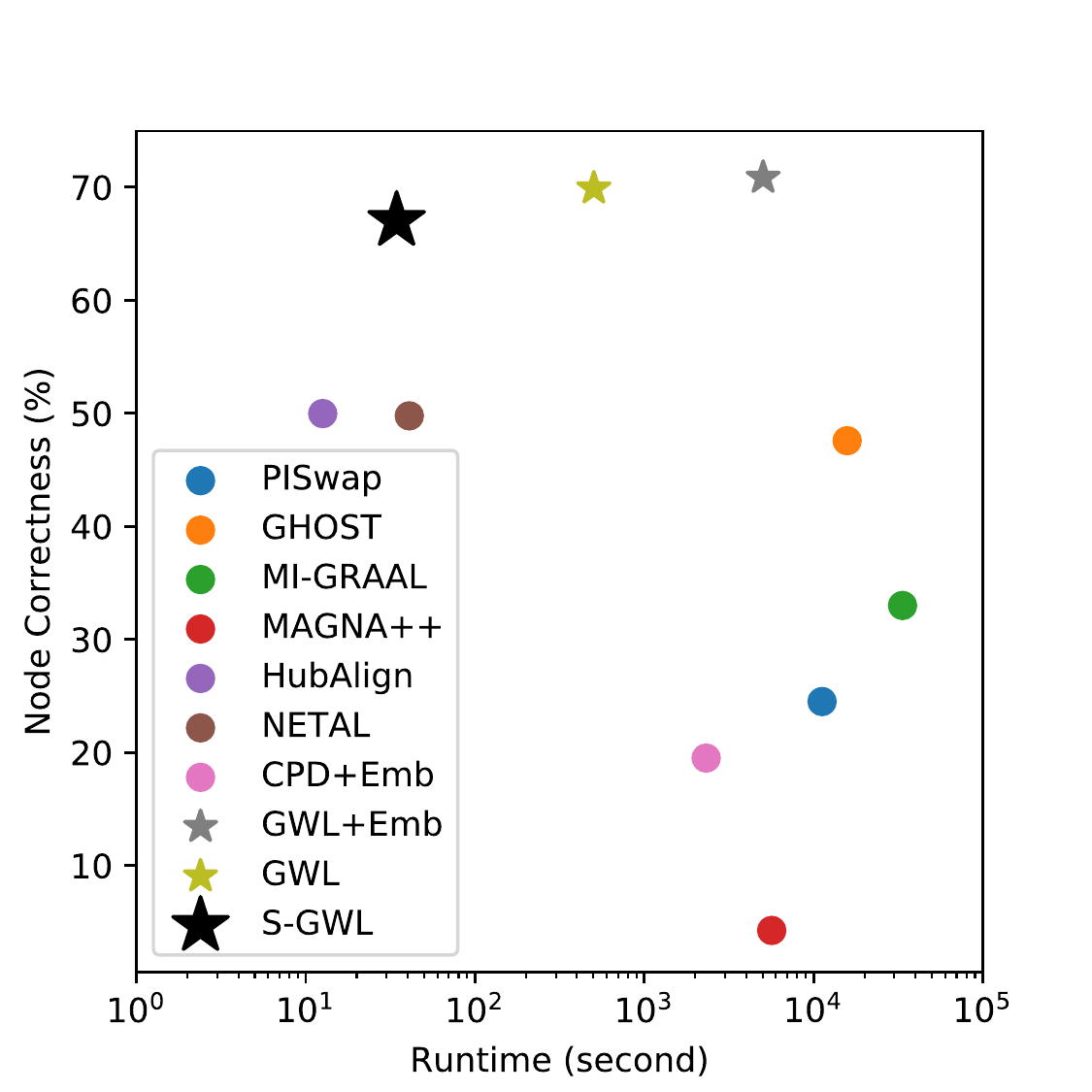}\label{fig:complexity}
    }
    \vspace{-9pt}
    \caption{\small{(a)-(d) Illustrations of graph partitioning and matching in the GWL framework. 
    (c, d) The barycenter graph in black and its optimal transports to observed graphs are learned jointly. 
    (d) When the barycenter graph is initialized as a graph with few isolated nodes, the optimal transports indicate aligned partitions of observed graph.
    (e) We test various graph matching methods in 10 trials on an Intel i7 CPU. 
    In each trial, the source graph has 2,000 nodes and the target graph has 100 more noisy nodes and corresponding edges. 
    The graphs yield either Gaussian partition model~\cite{brandes2003experiments} or  Barab{\'a}si-Albert model~\cite{barabasi2016network}. 
    The GWL-based methods (`$\bm{\star}$') obtains higher node correctness than other baselines (`$\bullet$'), and our S-GWL (big `$\bm{\star}$') achieves a trade-off on accuracy and efficiency. 
    }}\label{fig1}
    \vspace{-10pt}
\end{figure}

Focusing on the issues above, we design a scalable Gromov-Wasserstein learning (S-GWL) method and establish a new and unified paradigm for large-scale graph partitioning and matching. 
As illustrated in Figure~\ref{fig:gm}, given two graphs, the optimal transport associated with their Gromov-Wasserstein discrepancy provides the correspondence between their nodes. 
Similarly, graph partitioning corresponds to calculating the Gromov-Wasserstein discrepancy between an observed graph and a  disconnected graph, as shown in Figure~\ref{fig:gp}. 
The optimal transport connects each node of the observed graph with an isolated node of the disconnected graph, yielding a partitioning. 
In Figures~\ref{fig:gms} and~\ref{fig:gps}, taking advantage of the Gromov-Wasserstein barycenter in~\cite{peyre2016gromov}, we achieve multi-graph matching and partitioning by learning a ``barycenter graph''. 
% (the barycenter graph plays the role of the disconnected graph). 
For arbitrary two or more graphs, the correspondence (or the clustering structure) among their nodes can be established indirectly through their optimal transports to the barycenter graph.

The four tasks in Figures~\ref{fig:gm}-\ref{fig:gps} are explicitly unified in our Gromov-Wasserstein learning (GWL) framework, which corresponds to the same GW discrepancy-based optimization problem. 
To improve its scalability, we introduce a recursive mechanism to the GWL framework, which recursively applies $K$-way partitioning to decompose large graphs into a set of aligned sub-graph pairs, and then matches each pair of sub-graphs. 
When calculating GW discrepancy, we design a regularized proximal gradient method, that considers the prior information of nodes and performs updates by solving a series of convex sub-problems. 
The sparsity of edges further helps us reduce computations. 
These acceleration strategies yield our S-GWL method: for graphs with $V$ nodes and $E$ edges, its time complexity is $\mathcal{O}(K(E+V)\log_K V)$ and memory complexity is $\mathcal{O}(E+VK)$. 
To our knowledge, our S-GWL is the first to make GW discrepancy applicable to large-scale graph analysis. 
Figure~\ref{fig:complexity} illustrates the effectiveness of S-GWL on graph matching, with more results presented in Section~\ref{sec:exp}. 

\vspace{-6pt}
\section{Graph Analysis Based on Gromov-Wasserstein Learning}
\vspace{-8pt}
Denote a \emph{measure graph} as $G(\mathcal{V}, \bm{C}, \bm{\mu})$, where $\mathcal{V}=\{v_i\}_{i=1}^{|\mathcal{V}|}$ is the set of nodes, $\bm{C}=[c_{ij}]\in\mathbb{R}^{|\mathcal{V}|\times|\mathcal{V}|}$ is the adjacency matrix, and $\bm{\mu}=[\mu_i]\in\Sigma^{|\mathcal{V}|}$ is a Borel probability measure defined on $\mathcal{V}$. 
\textcolor{black}{The adjacency matrix is continuous for weighted graph while binary for unweighted graph.}
In practice, $\bm{\mu}$ is an empirical distribution of nodes, which can be estimated by \textcolor{black}{a function of node degree.} 
A $K$-way graph partitioning aims to decompose a graph $G$ into $K$ sub-graphs by clustering its nodes, $i.e.$, $\{G_k=G(\mathcal{V}_k, \bm{C}_k, \bm{\mu}_k)\}_{k=1}^{K}$, where $\cup_{k}\mathcal{V}_k=\mathcal{V}$ and $\mathcal{V}_k\cap\mathcal{V}_{k'}=\emptyset$ for $k\neq k'$. 
Given two graphs $G_s$ and $G_t$, graph matching aims to find a correspondence between their nodes, $i.e.$, $\pi:~\mathcal{V}_s\mapsto\mathcal{V}_t$. 
Many real-world networks are modeled using graph theory, and graph partitioning and matching are important for community detection~\cite{karypis1998fast,girvan2002community} and network alignment~\cite{sharan2006modeling,singh2008global,zhang2015multiple}, respectively.
In this section, we propose a Gromov-Wasserstein learning framework to unify these two problems.

\vspace{-6pt}
\subsection{Gromov-Wasserstein discrepancy between graphs}\label{sec:gwl}
\vspace{-8pt}
Our GWL framework is based on a pseudometric on graphs called Gromov-Wasserstein discrepancy:
\begin{definition}[\cite{chowdhury2018gromov}]\label{thm1}
Denote the collection of measure graphs as $\mathcal{G}$.
For each $p\in [1,\infty]$ and each $G_s,G_t\in\mathcal{G}$, the Gromov-Wasserstein discrepancy between $G_s$ and $G_t$ is
\vspace{-5pt}
\begin{eqnarray}\label{eq:ggwd}
d_{gw}(G_s, G_t):=\sideset{}{_{\bm{T}\in \Pi(\bm{\mu}_s,\bm{\mu}_t)}}\min\Bigl(
\sideset{}{_{i,j\in\mathcal{V}_s}}\sum \sideset{}{_{i',j'\in\mathcal{V}_t}}\sum |c_{ij}^s - c_{i'j'}^t|^{p}T_{ii'}T_{jj'}
\Bigr)^{\frac{1}{p}},
\end{eqnarray}
\vspace{-5pt}
where $\Pi(\bm{\mu}_s,\bm{\mu}_t)=\{\bm{T}\geq \bm{0}|\bm{T}\bm{1}_{|\mathcal{V}_t|}=\bm{\mu}_s,\bm{T}^{\top}\bm{1}_{|\mathcal{V}_s|}=\bm{\mu}_t\}$. 
\end{definition}
GW discrepancy compares graphs in a relational way, measuring how the edges in a graph compare to those in the other graph. 
It is a natural extension of the Gromov-Wasserstein distance defined for metric-measure spaces~\cite{memoli2011gromov}. 
We refer the reader to~\cite{memoli2011gromov,chowdhury2018gromov,peyre2019computational} for mathematical foundations.

\textbf{Graph matching} According to the definition, GW discrepancy measures the distance between two graphs, and the optimal transport $\bm{T}=[T_{ij}]\in\Pi(\bm{\mu}_s,\bm{\mu}_t)$ is a joint distribution of the graphs' nodes: $T_{ij}$ indicates the probability that the node $v_i^s\in\mathcal{V}_s$ corresponds to the node $v_j^t\in\mathcal{V}_t$. 
As shown in Figure~\ref{fig:gm}, the optimal transport achieves an assignment of the source nodes to the target ones. 

\textbf{Graph partitioning} Besides graph matching, this paradigm is also suitable for graph partitioning.
Recall that most existing graph partitioning methods obey the modularity maximization principle~\cite{girvan2002community,clauset2004finding}:  for each partitioned sub-graph, its internal edges should be dense, while its external edges connecting with other sub-graphs should be sparse. 
This principle implies that if we treat each sub-graph as a ``super node''~\cite{karypis1998fast,wang2011detecting,paresfluid}, an ideal partitioning should correspond to a disconnected graph with $K$ isolated, but self-connected super nodes. 
Therefore, we achieve $K$-way partitioning by calculating the GW discrepancy between the observed graph $G$ and a disconnected graph, % (defined in the next subsection in terms of a barycenter graph), 
$i.e.$, $d_{gw}(G,G_{\text{dc}})$, where $G_{\text{dc}}=G(\mathcal{V}_{\text{dc}}, \text{diag}(\bm{\mu}_{\text{dc}}),\bm{\mu}_{\text{dc}})$. 
$|\mathcal{V}_{\text{dc}}|=K$. 
$\bm{\mu}_{\text{dc}}\in\Sigma^{K}$ is a node distribution, whose derivation is in Appendix~\ref{apx1}. 
$\text{diag}(\bm{\mu}_{\text{dc}})$ is the adjacency matrix of $G_{\text{dc}}$.
As shown in Figure~\ref{fig:gp}, the optimal transport is a $|\mathcal{V}|\times K$ matrix. 
The maximum in each row of the matrix indicates the cluster of a node. 

\vspace{-6pt}
\subsection{Gromov-Wasserstein barycenter graph for analysis of multiple graphs}
\vspace{-8pt}
\textcolor{black}{\textbf{Multi-graph matching}} Distinct from most graph matching methods~\cite{gold1996graduated,cordella2004sub,sharan2006modeling,cour2007balanced}, which mainly focus on two-graph matching, our GWL framework can be readily extended to multi-graph cases, by introducing the Gromov-Wasserstein barycenter (GWB) proposed in~\cite{peyre2016gromov}.
Given a set of graphs $\{G_m\}_{m=1}^{M}$, their $p$-order Gromov-Wasserstein barycenter is a \emph{barycenter graph} defined as
\begin{eqnarray}\label{eq:gwb}
\begin{aligned}
G(\bar{\mathcal{V}}, \bar{\bm{C}}, \bar{\bm{\mu}}):=\arg\sideset{}{_{\bar{G}}}\min \sideset{}{_{m=1}^{M}}\sum \omega_{m}d_{gw}^p(G_m, \bar{G}),
\end{aligned}
\end{eqnarray}
where $\bm{\omega}=[\omega_m]\in\Sigma^{M}$ contains predefined weights, and $\bar{G}=G(\bar{\mathcal{V}}, \bar{\bm{C}}\in\mathbb{R}^{|\bar{\mathcal{V}}|\times |\bar{\mathcal{V}}|}, \bar{\bm{\mu}}\in\Sigma^{|\bar{\mathcal{V}}|})$ is the barycenter graph with a predefined number of nodes.
The barycenter graph minimizes the weighted average of its GW discrepancy to observed graphs. 
\textcolor{black}{It is an average of the observed graphs aligned by their optimal transports. 
The matrix $\bar{\bm{C}}$ is a ``soft'' adjacency matrix of the barycenter. 
Its elements reflect the confidence of the edges between the corresponding nodes in $\bar{\mathcal{V}}$.}
As shown in Figure~\ref{fig:gms}, the barycenter graph works as a ``reference'' connecting with the observed graphs. 
For each node in the barycenter graph, we can find its matched nodes in different graphs with the help of the corresponding optimal transport. 
These matched nodes construct a node set, and two arbitrary nodes in the set are a correspondence.
The collection of all the node sets achieves multi-graph matching. 

\textcolor{black}{\textbf{Multi-graph partitioning}} We can also use the barycenter graph to achieve multi-graph partitioning, with the \emph{learned} barycenter graph playing the role of the aforementioned disconnected graph. 
Given two or more graphs, whose nodes may have unobserved correspondences, existing partitioning methods~\cite{karypis1998fast,girvan2002community,clauset2004finding,blondel2008fast,paresfluid} only partition them independently because they are designed for clustering nodes in a single graph. 
As a result, the first cluster of a graph may correspond to the second cluster of another graph. 
Without the correspondence between clusters, we cannot reduce the search space in matching tasks. 
Although this correspondence can be estimated by matching two coarse graphs that treat the clusters as their nodes, this strategy not only introduces additional computations but also leads to more uncertainty on matching, because different graphs are partitioned independently without leveraging structural information from each other. 
By learning a barycenter graph for multiple graphs, we can partition them and align their clusters simultaneously.
As shown in Figure~\ref{fig:gps}, when applying $K$-way multi-graph partitioning, we initialize a disconnected graph with $K$ isolated nodes as the barycenter graph, and then learn it by $\min_{\bar{G}}\sum_{m=1}^{M} \omega_{m}d_{gw}^p(G_m, \bar{G})$. 
For each node of the barycenter graph, its matched nodes in each observed graph belong to the same cluster.

\vspace{-6pt}
\section{Scalable Gromov-Wasserstein Learning}\label{sec:sgwl}
\vspace{-8pt}
Based on Gromov-Wasserstein discrepancy and the barycenter graph, we have established a GWL framework for graph partitioning and matching. 
To make this framework scalable to large graphs, we propose a regularized proximal gradient method to calculate GW discrepancy and integrate multiple acceleration strategies to greatly reduce the computational complexity of GWL.

\vspace{-6pt}
\subsection{Regularized proximal gradient method}
\vspace{-8pt}
Inspired by the work in~\cite{xie2018fast,xu2019gromov}, we calculate the GW discrepancy in (\ref{eq:ggwd}) based on a proximal gradient method, which decomposes a complicated non-convex optimization problem into a series of convex sub-problems. 
For simplicity, we set $p=2$ in (\ref{eq:ggwd},~\ref{eq:gwb}). 
Given two graphs $G_s=G(\mathcal{V}_s, \bm{C}_s, \bm{\mu}_s)$ and $G_t=G(\mathcal{V}_t, \bm{C}_t, \bm{\mu}_t)$, in the $n$-th iteration, we update the current optimal transport $\bm{T}^{(n)}$ by calculating $d_{gw}^2(G_s, G_t)$:
\begin{eqnarray}\label{eq:proximal}
\begin{aligned}
\bm{T}^{(n+1)}&=\arg\sideset{}{_{\bm{T}\in \Pi(\bm{\mu}_s,\bm{\mu}_t)}}\min 
\sideset{}{_{i,j\in\mathcal{V}_s}}\sum \sideset{}{_{i',j'\in\mathcal{V}_t}}\sum |c_{ij}^s - c_{i'j'}^t|^{2}T_{ii'}^{(n)}T_{jj'}+\gamma \mbox{KL}(\bm{T}\lVert \bm{T}^{(n)})\\
&=\arg\sideset{}{_{\bm{T}\in \Pi(\bm{\mu}_s,\bm{\mu}_t)}}\min 
\langle \bm{L}(\bm{C}_s, \bm{C}_t,\bm{T}^{(n)}),\bm{T} \rangle + \gamma \mbox{KL}(\bm{T}\lVert \bm{T}^{(n)}).
% \\
% &=\arg\sideset{}{_{\bm{T}\in \Pi(\bm{\mu}_s,\bm{\mu}_t)}}\min 
% \langle \bm{L}(\bm{C}_s, \bm{C}_t,\bm{T}^{(n)}) - \gamma\log\bm{T}^{(n)},\bm{T} \rangle + \gamma \mbox{H}(\bm{T}).
\end{aligned}
\end{eqnarray}
Here, $\bm{L}(\bm{C}_s, \bm{C}_t, \bm{T})=\bm{C}_s\bm{\mu}_s\bm{1}_{|\mathcal{V}_t|}^{\top}+\bm{1}_{|\mathcal{V}_s|}\bm{\mu}_t^{\top}\bm{C}_t^{\top}-2\bm{C}_s\bm{T}\bm{C}_t^{\top}$, derived based on~\cite{peyre2016gromov}, and $\langle\cdot,\cdot\rangle$ represents the inner product of two matrices. 
The Kullback-Leibler (KL) divergence, $i.e.$, $\mbox{KL}(\bm{T}\lVert \bm{T}^{(n)})=\sum_{ij}T_{ij}\log({T_{ij}}/{T_{ij}^{(n)}}) - T_{ij} +T_{ij}^{(n)}$, is added as the proximal term.
% The third equality indicates that this can be reformulated as an optimal transport problem with an entropy regularizer~\cite{benamou2015iterative,peyre2016gromov}, where $\mbox{H}(\bm{T})=\sum_{ij}T_{ij}\log T_{ij}$. 
We can solve (\ref{eq:proximal}) via the Sinkhorn-Knopp algorithm~\cite{sinkhorn1967concerning,cuturi2013sinkhorn} with nearly-linear convergence~\cite{altschuler2017near}. 
As demonstrated in~\cite{xu2019gromov}, the global convergence of this proximal gradient method is guaranteed, so repeating (\ref{eq:proximal}) leads to a stable optimal transport, denoted as $\widehat{\bm{T}}$.
Additionally, this method is robust to hyperparameter $\gamma$, achieving better convergence and numerical stability than the entropy-based method in~\cite{peyre2016gromov}. 

Learning the barycenter graph is also based on the proximal gradient method. 
Given $M$ graphs, we estimate their barycenter graph via alternating optimization. 
In the $n$-th iteration, given the previous barycenter graph $\bar{G}^{(n)}=G(\bar{\mathcal{V}}, \bar{\bm{C}}^{(n)}, \bar{\bm{\mu}})$, we update $M$ optimal transports via solving (\ref{eq:proximal}).
Given the updated optimal transports $\{\bm{T}_m^{(n+1)}\}_{m=1}^{M}$, we update the adjacency matrix of the barycenter graph by
\begin{eqnarray}\label{eq:update}
%\small{
\begin{aligned}
% \bar{\bm{\mu}}^{(n+1)} = \sideset{}{_{m}}\sum \omega_m (\bm{T}_m^{(n+1)})^{\top} \bm{\mu}_m,~
\bar{\bm{C}}^{(n+1)} = \frac{1}{\bar{\bm{\mu}}\bar{\bm{\mu}}^{\top}}\sideset{}{_{m}}\sum \omega_m (\bm{T}_m^{(n+1)})^{\top} \bm{C}_m \bm{T}_m^{(n+1)}.
\end{aligned}
%}
\end{eqnarray}
The weights $\bm{\omega}$, the number of the nodes  $|\bar{\mathcal{V}}|$ and the node distribution $\bar{\bm{\mu}}$ are predefined.

Different from the work in~\cite{xu2019gromov,peyre2016gromov}, we use the following initialization strategies to achieve a regularized proximal gradient method and estimate optimal transports with few iterations.

\textcolor{black}{
\textbf{Node distributions} We estimate the node distribution $\bm{\mu}$ of a graph empirically by a function of node degree, which reflects the local topology of nodes, $e.g.$, the density of neighbors. 
In particular, for a graph with $|\mathcal{V}|$ nodes, we first calculate a vector of node degree, $i.e.$, $\bm{n}=[n_i]\in\mathbb{Z}^{|\mathcal{V}|}$, where $n_i$ is the number of neighbors of the $i$-th node.
Then, we estimate the node distribution $\bm{\mu}$ as
\begin{eqnarray}\label{eq:node}
\bm{\mu} = {\tilde{\bm{\mu}}}/{\|\tilde{\bm{\mu}}\|_1},\quad \tilde{\bm{\mu}} = (\bm{n} + a)^b.
\end{eqnarray}
where $a\geq 0$ and $b\geq 0$ are the hyperparameters controlling the shape of the distribution. 
For the graphs with isolated nodes, whose $n_i$'s are zeros, we set $a>0$ to avoid numerical issues when solving (\ref{eq:proximal}). 
For the graphs whose nodes obey to power-law distributions, $i.e.$, Barab{\'a}si-Albert graphs, we set $b\in [0, 1)$ to balance the probabilities of different nodes. 
This function generalizes the empirical settings used in other methods: when $a=0$ and $b=1$, we derive the distribution based on the normalized node degree used in~\cite{xu2019gromov}; when $b=0$, we assume the distribution is uniform as the work in~\cite{peyre2016gromov,vayer2018optimal} does. 
We find that the node distributions have a huge influence on the stability and the performance of our learning algorithms, which will be discussed in the following sections.
}

\textbf{Optimal transports}
For graph analysis, we can leverage prior knowledge to get a better regularization of optimal transport. 
Generally, the nodes with similar local topology should be matched with a high probability. 
Therefore, given two node distributions $\bm{\mu}_s$ and $\bm{\mu}_t$, we construct a node-based cost matrix $\bm{C}_{\text{node}}\in\mathbb{R}^{|\mathcal{V}_s|\times |\mathcal{V}_t|}$, whose element is $c_{ij} = |\mu_i^s - \mu_j^t|$, and add a regularization term $\langle \bm{C}_{\text{node}}, \bm{T}^{(n)}\rangle$ to (\ref{eq:proximal}). 
\textcolor{black}{As a result, in the learning phase, we replace the $\bm{L}(\bm{C}_s, \bm{C}_t,\bm{T}^{(n)})$ in (\ref{eq:proximal}) with $\bm{L}(\bm{C}_s, \bm{C}_t,\bm{T}^{(n)})+\tau\bm{C}_{\text{node}}$, where $\tau$ controls the significance of $\bm{C}_{\text{node}}$.} 
\textcolor{black}{Introducing the proposed regularizer helps us measure the similarity between nodes directly, which extends our GW discrepancy to the fused GW discrepancy in~\cite{vayer2018optimal,vayer2018fused}. 
In such a situation, the main difference here is that we use the proximal gradient method to calculate the discrepancy, rather than the conditional gradient method in~\cite{vayer2018fused}.}

\textbf{Barycenter graphs}
When learning GWB, the work in~\cite{peyre2016gromov} fixed the node distribution to be uniform 
In practice, however, both the node distribution of the barycenter graph and its optimal transports to observed graphs are unknown. 
In such a situation, we need to first estimate the node distribution $\bar{\bm{\mu}}=[\bar{\mu}_1,...,\bar{\mu}_{|\bar{\mathcal{V}}|}]$. 
Without loss of generality, we assume that the node distribution of the barycenter graph is sorted, $i.e.$, $\bar{\mu}_1\geq ...\geq\bar{\mu}_{|\bar{\mathcal{V}}|}$. 
We estimate the node distribution via the weighted average of the sorted and re-sampled node distributions of observed graphs:
\begin{eqnarray}\label{eq:wb}
\begin{aligned}
\bar{\bm{\mu}}=\sideset{}{_{m=1}^{M}}\sum \omega_m\text{interpolate}_{|\bar{\mathcal{V}}|}(\text{sort}(\bm{\mu}_{m})),
\end{aligned}
\end{eqnarray}
where $\text{sort}(\cdot)$ sorts the elements of the input vector in descending order, and $\text{interpolate}_{|\bar{\mathcal{V}}|}(\cdot)$ samples $|\bar{\mathcal{V}}|$ values from the input vector via bilinear interpolation. 
Given the node distribution, we initialize the optimal transports via the method mentioned above. 
\begin{figure*}[t]
\centering
\begin{minipage}[t]{0.53\linewidth}
\begin{algorithm}[H]
\small{
	\caption{$\text{ProxGrad}(G_s,G_t,\gamma)$}
	\label{alg:prox}
	\begin{algorithmic}[1]
	    \STATE Set $n=0$, $\bm{a}=\bm{\mu}_s$.
		\STATE Calculate $\bm{C}_{\text{node}}$ with $c_{ij}=|\mu_i^s-\mu_j^t|$. 
		\STATE Initialize $\bm{T}^{(n)} = \bm{\mu}_s\bm{\mu}_t^{\top}$.
		\STATE{\textbf{While} not converge}
		\STATE ~~$\bm{G}=e^{-(\bm{C}_{\text{node}}+\bm{L}(\bm{C}_s, \bm{C}_t,\bm{T}^{(n)}))/\gamma}\odot \bm{T}^{(n)}$.
        \STATE ~~$\bm{b}={\bm{\mu}_{t}}/{(\bm{G}^{\top}\bm{a})}$, and $\bm{a}={\bm{\mu}_{s}}/{(\bm{G}\bm{b})}$.
        \STATE ~~$\bm{T}^{(n+1)}=\text{diag}(\bm{a})\bm{G}\text{diag}(\bm{b})$, then $n=n+1$.
		\STATE \textbf{Output:} $\widehat{\bm{T}}=\bm{T}^{(n)}$.
	\end{algorithmic}
}
\end{algorithm}
\end{minipage}
\vspace{-8pt}
~~
\begin{minipage}[t]{0.44\linewidth}
\begin{algorithm}[H]
\small{
	\caption{$\text{GWB}(\{G_m\}_{m=1}^{M},\gamma,|\bar{\mathcal{V}}|,\bm{\omega})$}
	\label{alg:gwb}
	\begin{algorithmic}[1]
		\STATE Set $n=0$. 
		\STATE Initialize $\bar{\bm{\mu}}$ via (\ref{eq:wb}). $\bar{\bm{C}}^{(n)}=\text{diag}(\bar{\bm{\mu}})$.
		\STATE \textbf{While} not converge
		\STATE\quad \textbf{For} $m=1,...,M$
		\STATE\quad\quad $\bm{T}_m^{(n+1)}=\text{ProxGrad}(G_m,\bar{G}^{(n)},\gamma)$.
        \STATE\quad Calculate $\bar{\bm{C}}^{(n+1)}$ via (\ref{eq:update}).
        \STATE\quad $n=n+1$.
		\STATE \textbf{Output:} $\widehat{\bm{T}}_m=\bm{T}_m^{(n)}$ for $m=1,..,M$.
	\end{algorithmic}
}
\end{algorithm}
\end{minipage}
\vspace{-8pt}
\end{figure*}

Algorithms~\ref{alg:prox} and~\ref{alg:gwb} show the details of our method, where ``$\odot$'' and ``${\cdot}/{\cdot}$'' represent elementwise multiplication and division, respectively.
The GWL framework for the tasks in Figures~\ref{fig:gm}-\ref{fig:gps} are implemented based on these two algorithms, with details in Appendix~\ref{apx1}.

\vspace{-6pt}
\subsection{A recursive $K$-partition mechanism for large-scale graph matching}
\vspace{-8pt}
Assume that the observed graphs have comparable size, whose number of nodes and edges are denoted as $V$ and $E$, respectively. 
When using the proximal gradient method directly to calculate the GW discrepancy between two graphs, the time complexity, in the worst case, is $\mathcal{O}(V^3)$ because the $\bm{L}(\bm{C}_s,\bm{C}_t,\bm{T}^{(n)})$ in (\ref{eq:proximal}) involves $\bm{C}_s\bm{T}\bm{C}_t^{\top}$. 
Even if we consider the sparsity of edges and implement sparse matrix multiplications, the time complexity is still as high as $\mathcal{O}(EV)$. 

\begin{figure}[t]
    \centering
    \subfigure[Scheme of our S-GWL method]{
    \includegraphics[width=0.72\linewidth]{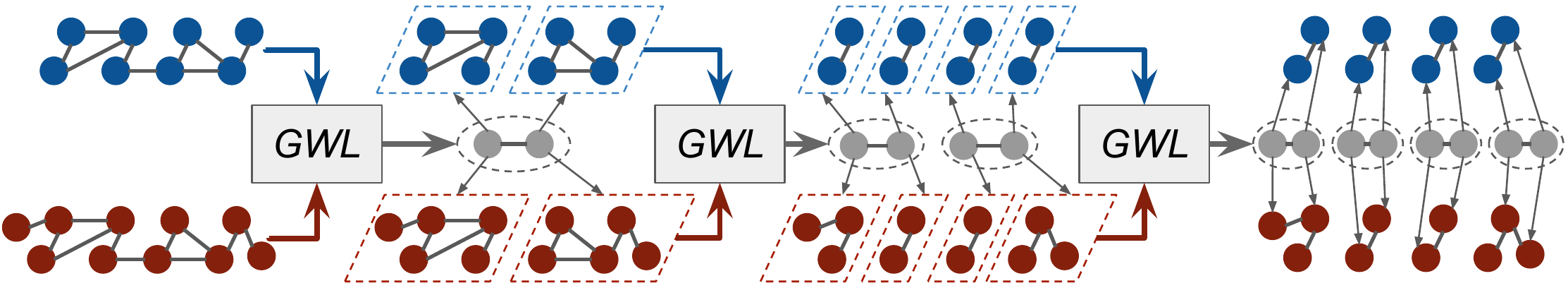}\label{fig:scheme}
    }
    \subfigure[Runtime]{
    \includegraphics[width=0.22\linewidth, height=0.14\linewidth]{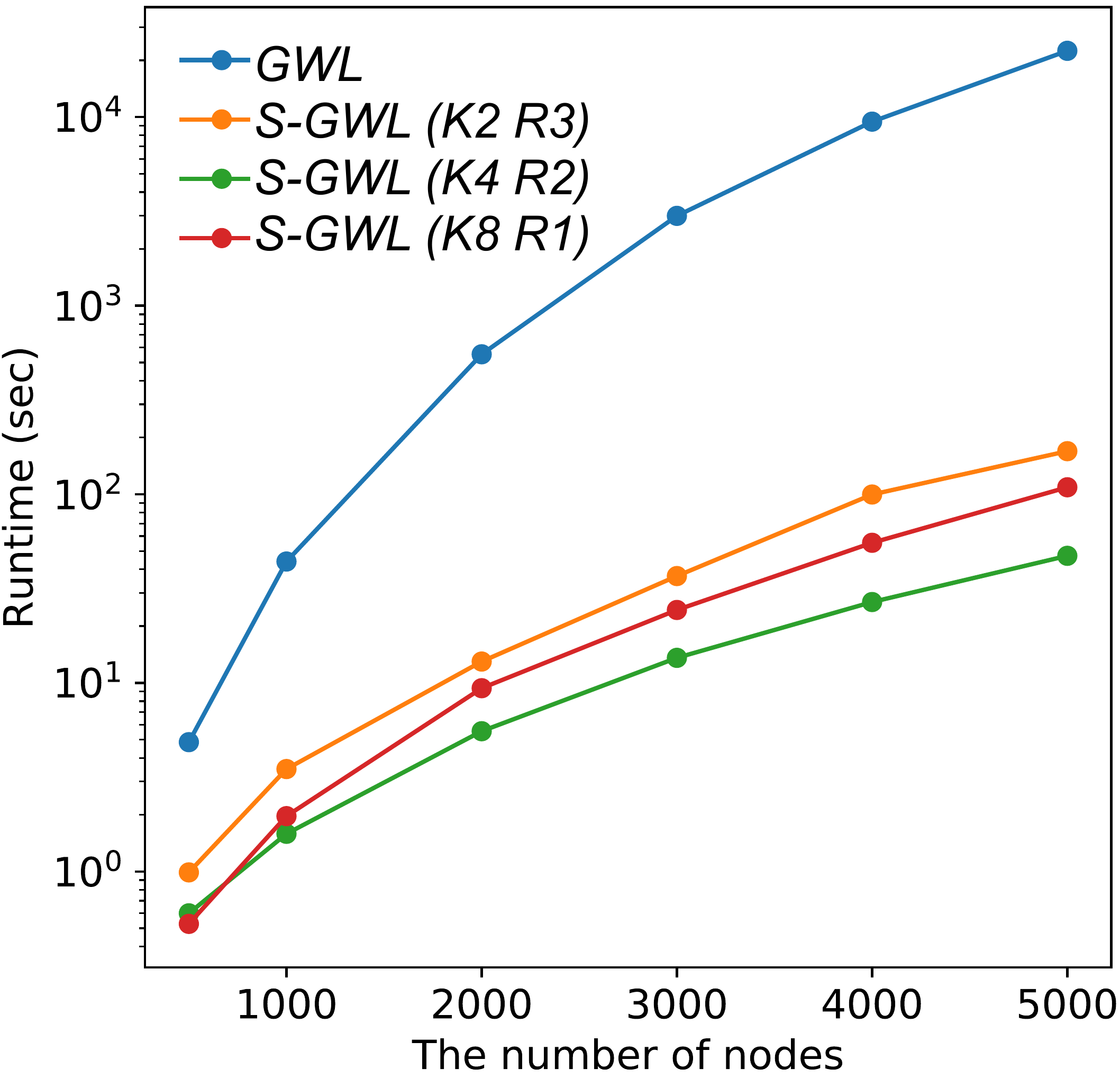}\label{fig:accelerate}
    }
    \vspace{-10pt}
    \caption{\small{(a) An illustration of S-GWL. (b) Comparisons on runtime.}}
    \vspace{-8pt}
\end{figure}

To improve the scalability of our GWL framework, we introduce a recursive $K$-partition mechanism, recursively decomposing observed large graphs to a set of aligned small graphs. 
As shown in Figure~\ref{fig:scheme}, given two graphs, we first calculate their barycenter graph (with $K$ nodes) and achieve their joint $K$-way partitioning. 
For each node of the barycenter graph, the corresponding sub-graphs extracted from the observed two graphs construct an aligned sub-graph pair, shown as the dotted frames connected with grey circles in Figure~\ref{fig:scheme}.
For each aligned sub-graph pair, we further calculate its barycenter graph and decompose the pair into more and smaller sub-graph pairs. 
Repeating the above step, we finally calculate the GW discrepancy between the sub-graphs in each pair, and find the correspondence between their nodes. 
Note that this recursive mechanism is also applicable to multi-graph matching: for multiple graphs, in the final step we calculate the GWB among the sub-graphs in each set.
The details of our S-GWL method are provided in Appendix~\ref{apx2}. 

\textbf{Complexity analysis}
In Table~\ref{tab:complex}, we compare the time and memory complexity of our S-GWL method with other matching methods.
The Hungarian algorithm~\cite{kuhn1955hungarian} has time complexity $\mathcal{O}(V^3)$~\cite{gold1996graduated,pachauri2013solving,yan2015consistency}. 
Denoting the largest node degree in a graph as $d$, the time complexity of GHOST~\cite{patro2012global} is $\mathcal{O}(d^4)$. 
The methods above take the graph affinity matrix as input, so their memory complexity in the worst case is $\mathcal{O}(V^4)$. 
MI-GRAAL~\cite{kuchaiev2011integrative}, HubAlign~\cite{hashemifar2014hubalign} and NETAL~\cite{neyshabur2013netal} are relatively efficient, with time complexity $\mathcal{O}(VE+V^2\log V)$, $\mathcal{O}(V^2\log V)$ and $\mathcal{O}(E^2+EV\log V)$, respectively. 
CPD+Emb first learns $D$-dimensional node embeddings~\cite{grover2016node2vec}, and then registers the embeddings by the CPD method~\cite{myronenko2010point}, whose time complexity is $\mathcal{O}(DV^2)$.  
The memory complexity of these four methods is $\mathcal{O}(V^2)$. 
For GW discrepancy-based methods, the GWL+Emb in~\cite{xu2019gromov} achieves graph matching and node embedding jointly. 
It uses the distance matrix of node embeddings and breaks the sparsity of edges, so its time complexity is $\mathcal{O}(V^3)$ and memory complexity is $\mathcal{O}(V^2)$. 
The time complexity of GWL is $\mathcal{O}(VE)$, but its memory complexity is still $\mathcal{O}(V^2)$ because the $\bm{L}(\bm{C}_s,\bm{C}_t,\bm{T}^{(n)})$ in (\ref{eq:proximal}) is a dense matrix. 
Our S-GWL combines the recursive mechanism with the regularized proximal gradient method and implements the $\bm{C}_s\bm{T}^{(n)}\bm{C}_t^{\top}$ in (\ref{eq:proximal}) by sparse matrix multiplications. 
Ideally, we can apply $R=\lfloor\log_K V\rfloor$ recursions. 
In the $r$-th recursion we calculate $K^r$ barycenter graphs for $K^r$ sub-graph pairs. 
The sub-graphs in each pair have $\mathcal{O}(\frac{V}{K^r})$ nodes. 
As a result, we have
\begin{proposition}\label{prop:complexity}
Suppose that we have $M$ graphs, each of which has $V$ nodes and $E$ edges. 
With the help of the recursive $K$-partition mechanism, 
the time complexity of our S-GWL method is $\mathcal{O}(MK(E+V)\log_{K}V)$, and its memory complexity is $\mathcal{O}(M(E+VK))$.
\end{proposition}
Choosing $K=2$ and ignoring the number of graphs, we obtain the complexity shown in Table~\ref{tab:complex}.
Our S-GWL has lower computational time complexity and memory requirements than many existing methods. 
Figure~\ref{fig:accelerate} visualizes the runtime of GWL and S-GWL on matching synthetic graphs. 
The S-GWL methods with different configurations ($i.e.$, the number of partitions $K$ and that of recursions $R$) are consistently faster than GWL. 
More detailed analysis is provided in Appendix~\ref{apx3}.

\begin{table}[!t]
\caption{Comparisons for graph matching methods on time and memory complexity.}
\vspace{-6pt}
\centering
\begin{threeparttable}
\footnotesize{
\setlength{\tabcolsep}{5pt}
\begin{tabular}{
@{\hspace{0pt}}c@{\hspace{2pt}}|
@{\hspace{2pt}}c@{\hspace{4pt}}c@{\hspace{4pt}}c@{\hspace{4pt}}c@{\hspace{4pt}}c@{\hspace{4pt}}c@{\hspace{2pt}}|
@{\hspace{2pt}}c@{\hspace{4pt}}c@{\hspace{4pt}}c@{\hspace{0pt}}
} 
\hline\hline 
%{Method}
&\tiny{Hungarian} 
&\tiny{GHOST$^*$} 
&\tiny{MI-GRAAL} 
&\tiny{HubAlign} 
&\tiny{NETAL} 
&\tiny{CPD+Emb} 
&\tiny{GWL+Emb}
&\tiny{GWL} 
&\tiny{S-GWL}\\\hline
Time $\mathcal{O}(\cdot)$
&$V^3$ 
&$d^4$
&$VE$+$V^2\log V$ 
&$V^2\log V$ 
&$E^2$+$EV\log V$ 
&$DV^2$ 
&$V^3$
&$VE$ 
&$2(E$+$V)\log V$\\
Memory $\mathcal{O}(\cdot)$
&$V^4$ 
&$V^4$ 
&$V^2$ 
&$V^2$ 
&$V^2$ 
&$V^2$ 
&$V^2$ 
&$V^2$
&$E+2V$\\
\hline\hline
\end{tabular}\label{tab:complex}
% \vspace{-2pt}
% \begin{tablenotes}
%  \item\tiny{To save space, we ignore the complexity symbol $\mathcal{O}(\cdot)$ and the number of graphs $M$ for all the methods.}
% \end{tablenotes}
}
\begin{tablenotes}
  \tiny{
  \item[*] $d$ is the largest node degree in a graph.
  }
\end{tablenotes}
\vspace{-8pt}
\end{threeparttable}
\end{table}

\vspace{-6pt}
\section{Related Work}
\vspace{-8pt}
\textbf{Gromov-Wasserstein learning}
GW discrepancy has been applied in many matching problems, $e.g.$, registering 3D objects~\cite{memoli2009spectral,memoli2011gromov} and matching vocabulary sets between different languages~\cite{alvarez2018gromov}. 
Focusing on graphs, a fused Gromov-Wasserstein distance is proposed  in~\cite{vayer2018optimal,vayer2018fused}, combining GW discrepancy with Wasserstein discrepancy~\cite{villani2008optimal}. 
The work in~\cite{xu2019gromov} further takes node embedding into account, learning the GW discrepancy between two graphs and their node embeddings jointly. 
The appropriateness of these methods is supported by~\cite{chowdhury2018gromov}, which proves that GW discrepancy is a pseudometric on measure graphs. 
Recently, an adversarial learning method based on GW discrepancy is proposed in~\cite{bunne2018}, which jointly trains two generative models in incomparable spaces. 
The work in~\cite{peyre2016gromov} further proposes Gromov-Wasserstein barycenters for clustering distributions and interpolating shapes. 
Currently, GW discrepancy is mainly calculated based on Sinkhorn iterations~\cite{sinkhorn1967concerning,cuturi2013sinkhorn,benamou2015iterative,peyre2016gromov}, whose applications to large-scale graphs are challenging because of its high complexity.
Our S-GWL method is the first attempt to make GW discrepancy applicable to large-scale graph analysis.

\textbf{Graph partitioning and graph matching}
Graph partitioning is important for community detection in networks. 
Many graph partitioning methods have been proposed, such as Metis~\cite{karypis1998fast}, EdgeBetweenness~\cite{girvan2002community}, FastGreedy~\cite{clauset2004finding}, Label Propagation~\cite{raghavan2007near}, Louvain~\cite{blondel2008fast} and Fluid Community~\cite{paresfluid}. 
All of these methods explore the clustering structure of nodes heuristically based on the modularity-maximization principle~\cite{girvan2002community,clauset2004finding}. 
Graph matching is important for network alignment~\cite{sharan2006modeling,singh2008global,zhang2015multiple} and 2D/3D object registration~\cite{myronenko2010point,yan2015matrix,jun2017sequential,yu2018generalizing}.
Traditional methods formulate graph matching as a quadratic assignment problem (QAP) and solve it based on the Hungarian algorithm~\cite{gold1996graduated,pachauri2013solving,yan2015matrix,yan2015consistency}, which are only applicable to small graphs.
For large graphs like protein networks, many heuristic methods have been proposed, such as GRAAL~\cite{kuchaiev2010topological}, IsoRank~\cite{singh2008global}, PISwap~\cite{chindelevitch2013optimizing},  MAGNA++~\cite{vijayan2015magna++}, NETAL~\cite{neyshabur2013netal}, HubAlign~\cite{hashemifar2014hubalign}, and GHOST~\cite{patro2012global}, which mainly focus on two-graph matching and are sensitive to the noise in graphs. 
With the help of GW discrepancy, our work establishes a unified framework for graph partitioning and matching, that can be readily extended to multi-graph cases.

\vspace{-6pt}
\section{Experiments\label{sec:exp}}
\vspace{-8pt}
\textcolor{black}{The implementation of our S-GWL method can be found at \url{https://github.com/HongtengXu/s-gwl}.}
We compare it with state-of-the-art methods for graph partitioning and matching. 
All the methods are run on an Intel i7 CPU with 4GB memory. 
Implementation details and a further set of experimental results are provided in Appendix~\ref{apx4}.

\vspace{-6pt}
\subsection{Graph partitioning}
\vspace{-8pt}
We first verify the performance of the $\textbf{GWL}$ framework on graph partitioning, comparing it with the following four baselines: \textbf{Metis}~\cite{karypis1998fast}, \textbf{FastGreedy}~\cite{clauset2004finding}, \textbf{Louvain}~\cite{blondel2008fast}, and \textbf{Fluid} Community~\cite{paresfluid}. 
We consider synthetic and real-world data.
Similar to~\cite{yang2016comparative}, we compare these methods in terms of adjusted mutual information (\href{https://en.wikipedia.org/wiki/Adjusted_mutual_information}{AMI}) and runtime. 
Each synthetic graph is a Gaussian random partition graph with $N$ nodes and $K$ clusters. 
The size of each cluster is drawn from a normal distribution $\mathcal{N}(200, 10)$. 
The nodes are connected within clusters with probability $p_{\text{in}}$ and between clusters with probability $p_{\text{out}}$. 
The ratio $\frac{p_{\text{out}}}{p_{\text{in}}}$ indicates the clearness of the clustering structure, and accordingly the difficulty of partitioning. 
We set $N=4000$, $p_{\text{in}}=0.2$, and $p_{\text{out}}\in \{0.05, 0.1, 0.15\}$.
Under each configuration $(N, p_{\text{in}}, p_{\text{out}})$, we simulate $10$ graphs. 
For each method, its average performance on these $10$ graphs is listed in Table~\ref{tab:syn_gp}.
GWL outperforms the alternatives consistently on AMI. 
Additionally, as shown in Table~\ref{tab:syn_gp}, GWL has time complexity comparable to other methods, especially when the graph is sparse, $e.g.$, $E=\mathcal{O}(V\log V)$. 
According to the runtime in practice, GWL is faster than most baselines except Metis, likely because Metis is implemented in the C language while GWL and other methods are based on Python. 
\begin{table}[!t]
\caption{Comparisons for graph partitioning methods on AMI, time complexity and runtime (second).}
\vspace{-6pt}
\centering
\small{
\setlength{\tabcolsep}{5pt}
\begin{tabular}{@{\hspace{0pt}}c|
c@{\hspace{3pt}}c|
c@{\hspace{3pt}}c|
c@{\hspace{3pt}}c|
c@{\hspace{3pt}}c|
c@{\hspace{3pt}}c@{\hspace{0pt}}
} 
\hline\hline
Method &
\multicolumn{2}{c|}{Metis} &
\multicolumn{2}{c|}{FastGreedy} &
\multicolumn{2}{c|}{Louvain} &
\multicolumn{2}{c|}{Fluid} &
\multicolumn{2}{c}{GWL}\\ \hline
Time complexity&
\multicolumn{2}{c|}{$\mathcal{O}(V$+$E$+$K\log K)$} &
\multicolumn{2}{c|}{$\mathcal{O}(VE\log V)$} &
\multicolumn{2}{c|}{$\mathcal{O}(V\log V)$} &
\multicolumn{2}{c|}{$\mathcal{O}(E)$} &
\multicolumn{2}{c}{$\mathcal{O}((E+V)K)$} \\ \hline
$(N,p_{\text{in}},p_{\text{out}})$&
AMI &Time &
AMI &Time &
AMI &Time &
AMI &Time &
AMI &Time \\ \hline
$(4000,0.2,0.05)$ &
0.413& 1.744&
0.247& 55.435&
0.747& 22.889&
0.776& 21.580&
\textbf{0.812}& 13.033\\
$(4000,0.2,0.1)$ &
0.009& 2.340&
0.064& 65.441&
0.574& 95.114&
0.577& 111.043&
\textbf{0.590}& 12.740\\
$(4000,0.2,0.15)$ &
0.002& 3.592&
0.002& 80.322&
0.005& 290.846&
0.005& 203.225&
\textbf{0.012}& 12.901\\
\hline\hline
\end{tabular}\label{tab:syn_gp}
}\vspace{-8pt}
\end{table}

\begin{table}[!t]
\caption{Comparisons for graph partitioning methods on AMI.}
\vspace{-6pt}
\centering
\begin{threeparttable}
\small{
\setlength{\tabcolsep}{5pt}
\begin{tabular}{c|cc|cc|cc|cc|cc
} 
\hline\hline
Method &
\multicolumn{2}{c|}{Metis} &
\multicolumn{2}{c|}{FastGreedy} &
\multicolumn{2}{c|}{Louvain} &
\multicolumn{2}{c|}{Fluid} &
\multicolumn{2}{c}{GWL}\\ \hline
Dataset &
Raw &Noisy &
Raw &Noisy &
Raw &Noisy &
Raw &Noisy &
Raw &Noisy \\ \hline
EU-Email &
0.421& 0.246&
0.312& 0.118&
0.434& 0.272&
---  & 0.338&
\textbf{0.459}& \textbf{0.349}\\
Indian-Village &
0.834& 0.513&
\textbf{0.882}& 0.275&
0.880& 0.633&
---  & 0.401&
0.857& \textbf{0.664}\\
\hline\hline
\end{tabular}\label{tab:real_gp}
\vspace{-2pt}
\begin{tablenotes}
 \item\tiny{``---'': Fluid is inapplicable when the networks have disconnected nodes or sub-graphs.}
\end{tablenotes}
}\vspace{-8pt}
\end{threeparttable}
\end{table}

Table~\ref{tab:real_gp} lists the performance of different methods on two real-world datasets. 
The first dataset is the email network from a large European research institution~\cite{snapnets}.
The network contains 1,005 nodes and 25,571 edges. 
The edge $(v_i, v_j)$ in the network mean that person $v_i$ sent person $v_j$ at least one email, and each node in the network belongs to exactly one of 42 departments at the research institute.
The second dataset is the interactions among 1,991 villagers in 12 Indian villages~\cite{banerjee2013diffusion}. 
Furthermore, to verify the robustness of GWL to noise, we not only consider the raw data of these two datasets but also create their noisy version by adding 10\% more noisy edges between different communities ($i.e.$, departments and villages). 
Experimental results show that GWL is at least comparable to its competitors on raw data, and it is more robust to noise than other methods.

\vspace{-6pt}
\subsection{Graph matching}
\vspace{-8pt}
For two-graph matching, we compare our S-GWL method with the following baselines:  \textbf{PISwap}~\cite{chindelevitch2013optimizing}, \textbf{GHOST}~\cite{patro2012global}, \textbf{MI-GRAAL}~\cite{kuchaiev2011integrative}, \textbf{MAGNA++}~\cite{vijayan2015magna++}, \textbf{HubAlign}~\cite{hashemifar2014hubalign}, \textbf{NETAL}~\cite{neyshabur2013netal},  \textbf{CPD+Emb}~\cite{grover2016node2vec,myronenko2010point}, the \textbf{GWL} framework based on Algorithm~\ref{alg:prox}, and the \textbf{GWL+Emb} in~\cite{xu2019gromov}. 
We test all methods on both synthetic and real-world data. 
For each method, given the learned correspondence set $\mathcal{S}$ and the ground-truth correspondence set $\mathcal{S}_{real}$, we calculate node correctness as $\text{NC}={|\mathcal{S}\cap\mathcal{S}_{real}|}/{|\mathcal{S}|}\times 100\%$. 
The runtime of each method is recorded as well.

In the synthetic dataset, each source graph $G(\mathcal{V}_s,\mathcal{E}_s)$ obeys a Gaussian random partition model~\cite{brandes2003experiments} or Barab{\'a}si-Albert model~\cite{barabasi2016network}. 
For each source graph, we generate a target graph by adding $|\mathcal{V}_s|\times q\%$ noisy nodes and $|\mathcal{E}_s|\times q\%$ noisy edges to the source graph. 
Figure~\ref{fig:complexity} compares our S-GWL with the baselines when $|\mathcal{V}_s|=2000$ and $q=5$. 
For each method, its average node correctness and runtime on matching 10 synthetic graph pairs are plotted. 
Compared with existing heursitic methods, GW discrepancy-based methods (GWL+Emb, GWL and S-GWL) obtain much higher node correctness. 
GWL+Emb achieves the highest node correctness, with runtime comparable to many baselines.
Our GWL framework does not learn node embeddings when matching graphs, so it is slightly worse than GWL+Emb on node correctness but achieves about 10 times acceleration.
Our S-GWL method further accelerates GWL with the help of the recursive mechanism. 
It obtains high node correctness and makes its runtime comparable to the fastest methods (HubAlign and NETAL).

In addition to graph matching on synthetic data, we also consider two real-world matching tasks.
The first task is matching the protein-protein interaction (PPI) network of yeast with its noisy version.
The PPI network of yeast contains 1,004 proteins and their 4,920 high-confidence interactions.
Its noisy version contains $q\%$ more low-confidence interactions, and $q\in\{5, 10, 15, 20, 25\}$.
The dataset is available on \url{https://www3.nd.edu/~cone/MAGNA++/}. 
The second task is matching user accounts in different communication networks.
The dataset is available on \url{http://vacommunity.org/VAST+Challenge+2018+MC3}, which records the communications among a company's employees. 
Following the work in~\cite{xu2019gromov}, we extract $622$ employees and their \emph{call-network} and \emph{email-network}. 
For each communication network, we construct a dense version and a sparse one: the dense version keeps all the communications (edges) among the employees, while the sparse version only preserves the communications happening more than $8$ times. 
We test different methods on $i$) matching yeast's PPI network with its $5\%$, $15\%$ and $25\%$ noisy versions; and $ii$) matching the employee call-network with their email-network in both sparse and dense cases. 
Table~\ref{tab:syn_gm} shows the performance of various methods in these two tasks. 
Similar to the experiments on synthetic data, the GW discrepancy-based methods outperform other methods on node correctness, especially for highly-noisy graphs, and our S-GWL method achieves a good trade-off between accuracy and efficiency.

\begin{table}[!t]
\caption{Comparisons for graph matching methods on node correctness (\%) and runtime (second).}
\vspace{-6pt}
\centering
\small{
\setlength{\tabcolsep}{5pt}
\begin{tabular}{c|cc|cc|cc|cc|cc
} 
\hline\hline
Dataset & 
\multicolumn{2}{c|}{Yeast 5\% noise} &
\multicolumn{2}{c|}{Yeast 15\% noise}&
\multicolumn{2}{c|}{Yeast 25\% noise}&
\multicolumn{2}{c|}{MC3 sparse}&
\multicolumn{2}{c}{MC3 dense}
\\ \hline
Method &
NC & Time &
NC & Time &
NC & Time &
NC & Time &
NC & Time
\\ \hline
PISwap &
0.10 & 15.80 &
0.10 & 18.31 &
0.00 & 22.09 &
6.32 & 10.27 &
0.00 & 11.81\\
GHOST &
11.06 & 25.67&
0.40  & 30.22&
0.30  & 35.54&
21.27 & 17.86&
0.03  & 22.90\\
MI-GRAAL &
18.03 & 189.21 &
6.87  & 202.77 &
5.18  & 240.03 &
35.53 & 72.89  &
0.64  & 197.65\\
MAGNA++ &
48.13 & 603.29 &
25.04 & 630.60 &
13.61 & 624.17 &
7.88  & 425.16 &
0.09  & 447.86 \\
HubAlign &
50.00 & 3.27 &
35.16 & 3.50 &
12.85 & 3.89 &
36.21 & 2.11 &
3.86  & 2.29 \\
NETAL &
6.87  & 1.91 &
0.90  & 2.06 &
1.00  & 2.09 &
36.87 & 1.23 &
1.77  & 1.30 \\
CPD+Emb &
3.59  & 103.22 &
2.09  & 110.19 &
2.00  & 108.62 &
4.35  & 87.54  &
0.48  & 95.68 \\ \hline
GWL+Emb &
83.66 & 1340.58 &
66.63 & 1499.20 &
57.97 & 1537.93 &
40.45 & 608.76  &
4.23  & 831.80 \\
GWL &
82.37 & 190.97 &
65.34 & 212.16 &
58.76 & 210.86 &
34.21 & 89.43 &
3.96  & 93.94 \\
S-GWL &
81.08 & 68.58 &
61.85 & 70.06 &
56.27 & 74.64 &
36.92 & 8.39  &
4.03  & 9.01  \\
\hline\hline
\end{tabular}\label{tab:syn_gm}
}\vspace{-8pt}
\end{table}

\begin{table}[!t]
\caption{Comparisons for multi-graph matching methods on yeast networks.}
\vspace{-6pt}
\centering
\small{
\setlength{\tabcolsep}{5pt}
\begin{tabular}{c|cc|cc|cc|cc
} 
\hline\hline
\multirow{2}{*}{Method} &
\multicolumn{2}{c|}{3 graphs} &
\multicolumn{2}{c|}{4 graphs} &
\multicolumn{2}{c|}{5 graphs} &
\multicolumn{2}{c}{6 graphs}\\ \cline{2-9}
&
NC@1 &NC@all &
NC@1 &NC@all &
NC@1 &NC@all &
NC@1 &NC@all \\ \hline
MultiAlign 
&62.97 &45.19
&--- &---
&--- &---
&--- &---\\
GWL
&\textbf{63.84} &\textbf{46.22}
&\textbf{68.73} &\textbf{39.14}
&71.61 &31.57
&76.49 &28.39\\
S-GWL
&60.06 &43.33
&68.53 &38.45
&\textbf{73.21} &\textbf{33.27}
&\textbf{76.99} &\textbf{29.68}\\
\hline\hline
\end{tabular}\label{tab:real_mgm}
}\vspace{-8pt}
\end{table}

Given the PPI network of yeast and its 5 noisy versions, we test GWL and S-GWL for multi-graph matching. 
We consider several existing multi-graph matching methods and find that the methods in~\cite{pachauri2013solving,yan2015matrix,yan2015consistency} are not applicable for the graphs with hundreds of nodes because $i$) their time complexity is at least $\mathcal{O}(V^3)$, and $ii$) they suffer from inadequate memory on our machine (with 4GB memory) because their memory complexity in the worst case is $\mathcal{O}(V^4)$. 
The IsoRankN in~\cite{liao2009isorankn} can align multiple PPI networks jointly, but it needs confidence scores of protein pairs as input, which are not available for our dataset. 
The only applicable baseline we are aware of is the \textbf{MultiAlign} in~\cite{zhang2015multiple}. 
However, it can only achieve three-graph matching. 
Table~\ref{tab:real_mgm} lists the performance of various methods.
Given learned correspondence sets, each of which is a set of matched nodes from different graphs, NC@1 represents the percentage of the set containing at least a pair of correctly-matched nodes, and NC@all represents the percentage of the set in which arbitrary two nodes are matched correctly. 
Both GWL and S-GWL obtain comparable performance to MultiAlign on three-graph matching, and GWL is the best. 
When the number of graphs increases, NC@1 increases while NC@all decreases for all the methods, and S-GWL becomes even better than GWL.

\vspace{-6pt}
\section{Conclusion and Future Work}
\vspace{-8pt}
We have developed a scalable Gromov-Wasserstein learning method, achieving large-scale graph partitioning and matching in a unified framework, with theoretical support. 
Experiments show that our approach outperforms state-of-the-art methods in many situations. 
\textcolor{black}{However, it should be noted that our S-GWL method is sensitive to its hyperparameters. 
Specifically, we observed in our experiments that the $\gamma$ in~(\ref{eq:proximal}) should be set carefully according to observed graphs.
Generally, for large-scale graphs we have to use a large $\gamma$  and solve (\ref{eq:proximal}) with many iterations. 
The $a$ and $b$ in (\ref{eq:node}) are also significant for the performance of our method. 
The settings of these hyperparameters and their influences are shown in Appendix~\ref{apx4}.}
In the future, we will further study the influence of hyperparameters on the rate of convergence and set the hyperparameters adaptively according to observed data.
Additionally, our S-GWL method can decompose a large graph into many independent small graphs, so we plan to further accelerate it by parallel processing and/or distributed learning. 

\textbf{Acknowledgements}
This research was supported in part by DARPA, DOE, NIH, ONR and NSF. 
We thank Dr. Hongyuan Zha for helpful discussions.

\bibliographystyle{ieee}
\bibliography{sgwl}

\begin{thebibliography}{10}\itemsep=-1pt

\bibitem{altschuler2017near}
J.~Altschuler, J.~Weed, and P.~Rigollet.
\newblock Near-linear time approximation algorithms for optimal transport via
  sinkhorn iteration.
\newblock In {\em Advances in Neural Information Processing Systems}, pages
  1964--1974, 2017.

\bibitem{alvarez2018gromov}
D.~Alvarez-Melis and T.~Jaakkola.
\newblock Gromov-wasserstein alignment of word embedding spaces.
\newblock In {\em Proceedings of the 2018 Conference on Empirical Methods in
  Natural Language Processing}, pages 1881--1890, 2018.

\bibitem{banerjee2013diffusion}
A.~Banerjee, A.~G. Chandrasekhar, E.~Duflo, and M.~O. Jackson.
\newblock The diffusion of microfinance.
\newblock {\em Science}, 341(6144):1236498, 2013.

\bibitem{barabasi2016network}
A.-L. Barab{\'a}si et~al.
\newblock {\em Network science}.
\newblock Cambridge university press, 2016.

\bibitem{benamou2015iterative}
J.-D. Benamou, G.~Carlier, M.~Cuturi, L.~Nenna, and G.~Peyr{\'e}.
\newblock Iterative {B}regman projections for regularized transportation
  problems.
\newblock {\em SIAM Journal on Scientific Computing}, 37(2):A1111--A1138, 2015.

\bibitem{blondel2008fast}
V.~D. Blondel, J.-L. Guillaume, R.~Lambiotte, and E.~Lefebvre.
\newblock Fast unfolding of communities in large networks.
\newblock {\em Journal of statistical mechanics: theory and experiment},
  2008(10):P10008, 2008.

\bibitem{brandes2003experiments}
U.~Brandes, M.~Gaertler, and D.~Wagner.
\newblock Experiments on graph clustering algorithms.
\newblock In {\em European Symposium on Algorithms}, pages 568--579. Springer,
  2003.

\bibitem{bronstein2010gromov}
A.~M. Bronstein, M.~M. Bronstein, R.~Kimmel, M.~Mahmoudi, and G.~Sapiro.
\newblock A {G}romov-{H}ausdorff framework with diffusion geometry for
  topologically-robust non-rigid shape matching.
\newblock {\em International Journal of Computer Vision}, 89(2-3):266--286,
  2010.

\bibitem{bunne2018}
C.~Bunne, D.~Alvarez-Melis, A.~Krause, and S.~Jegelka.
\newblock Learning generative models across incomparable spaces.
\newblock {\em NeurIPS Workshop on Relational Representation Learning}, 2018.

\bibitem{chindelevitch2013optimizing}
L.~Chindelevitch, C.-Y. Ma, C.-S. Liao, and B.~Berger.
\newblock Optimizing a global alignment of protein interaction networks.
\newblock {\em Bioinformatics}, 29(21):2765--2773, 2013.

\bibitem{chowdhury2018gromov}
S.~Chowdhury and F.~M{\'e}moli.
\newblock The {G}romov-{W}asserstein distance between networks and stable
  network invariants.
\newblock {\em arXiv preprint arXiv:1808.04337}, 2018.

\bibitem{clauset2004finding}
A.~Clauset, M.~E. Newman, and C.~Moore.
\newblock Finding community structure in very large networks.
\newblock {\em Physical review E}, 70(6):066111, 2004.

\bibitem{cordella2004sub}
L.~P. Cordella, P.~Foggia, C.~Sansone, and M.~Vento.
\newblock A (sub) graph isomorphism algorithm for matching large graphs.
\newblock {\em IEEE Transactions on Pattern Analysis and Machine Intelligence},
  26(10):1367--1372, 2004.

\bibitem{cour2007balanced}
T.~Cour, P.~Srinivasan, and J.~Shi.
\newblock Balanced graph matching.
\newblock In {\em NIPS}, pages 313--320, 2007.

\bibitem{cuturi2013sinkhorn}
M.~Cuturi.
\newblock Sinkhorn distances: Lightspeed computation of optimal transport.
\newblock In {\em Advances in neural information processing systems}, pages
  2292--2300, 2013.

\bibitem{girvan2002community}
M.~Girvan and M.~E. Newman.
\newblock Community structure in social and biological networks.
\newblock {\em Proceedings of the national academy of sciences},
  99(12):7821--7826, 2002.

\bibitem{gold1996graduated}
S.~Gold and A.~Rangarajan.
\newblock A graduated assignment algorithm for graph matching.
\newblock {\em IEEE Transactions on Pattern Analysis and Machine Intelligence},
  18(4):377--388, 1996.

\bibitem{grover2016node2vec}
A.~Grover and J.~Leskovec.
\newblock node2vec: {S}calable feature learning for networks.
\newblock In {\em KDD}, pages 855--864, 2016.

\bibitem{hashemifar2014hubalign}
S.~Hashemifar and J.~Xu.
\newblock Hubalign: {A}n accurate and efficient method for global alignment of
  protein--protein interaction networks.
\newblock {\em Bioinformatics}, 30(17):i438--i444, 2014.

\bibitem{jun2017sequential}
S.-H. Jun, S.~W. Wong, J.~Zidek, and A.~Bouchard-C{\^o}t{\'e}.
\newblock Sequential graph matching with sequential monte carlo.
\newblock In {\em AISTATS}, pages 1075--1084, 2017.

\bibitem{karypis1998fast}
G.~Karypis and V.~Kumar.
\newblock A fast and high quality multilevel scheme for partitioning irregular
  graphs.
\newblock {\em SIAM Journal on scientific Computing}, 20(1):359--392, 1998.

\bibitem{kuchaiev2010topological}
O.~Kuchaiev, T.~Milenkovi{\'c}, V.~Memi{\v{s}}evi{\'c}, W.~Hayes, and
  N.~Pr{\v{z}}ulj.
\newblock Topological network alignment uncovers biological function and
  phylogeny.
\newblock {\em Journal of the Royal Society Interface}, page rsif20100063,
  2010.

\bibitem{kuchaiev2011integrative}
O.~Kuchaiev and N.~Pr{\v{z}}ulj.
\newblock Integrative network alignment reveals large regions of global network
  similarity in yeast and human.
\newblock {\em Bioinformatics}, 27(10):1390--1396, 2011.

\bibitem{kuhn1955hungarian}
H.~W. Kuhn.
\newblock The hungarian method for the assignment problem.
\newblock {\em Naval research logistics quarterly}, 2(1-2):83--97, 1955.

\bibitem{snapnets}
J.~Leskovec and A.~Krevl.
\newblock {SNAP Datasets}: {Stanford} large network dataset collection.
\newblock \url{http://snap.stanford.edu/data}, June 2014.

\bibitem{liao2009isorankn}
C.-S. Liao, K.~Lu, M.~Baym, R.~Singh, and B.~Berger.
\newblock Isorankn: spectral methods for global alignment of multiple protein
  networks.
\newblock {\em Bioinformatics}, 25(12):i253--i258, 2009.

\bibitem{malod2015graal}
N.~Malod-Dognin and N.~Pr{\v{z}}ulj.
\newblock {L-GRAAL}: {L}agrangian graphlet-based network aligner.
\newblock {\em Bioinformatics}, 31(13):2182--2189, 2015.

\bibitem{memoli2009spectral}
F.~M{\'e}moli.
\newblock Spectral {G}romov-{W}asserstein distances for shape matching.
\newblock In {\em ICCV Workshops}, pages 256--263, 2009.

\bibitem{memoli2011gromov}
F.~M{\'e}moli.
\newblock Gromov-{W}asserstein distances and the metric approach to object
  matching.
\newblock {\em Foundations of computational mathematics}, 11(4):417--487, 2011.

\bibitem{memoli2004comparing}
F.~M{\'e}moli and G.~Sapiro.
\newblock Comparing point clouds.
\newblock In {\em Proceedings of the 2004 Eurographics/ACM SIGGRAPH symposium
  on Geometry processing}, pages 32--40, 2004.

\bibitem{myronenko2010point}
A.~Myronenko and X.~Song.
\newblock Point set registration: {C}oherent point drift.
\newblock {\em IEEE Transactions on Pattern Analysis and Machine Intelligence},
  32(12):2262--2275, 2010.

\bibitem{neyshabur2013netal}
B.~Neyshabur, A.~Khadem, S.~Hashemifar, and S.~S. Arab.
\newblock {NETAL}: {A} new graph-based method for global alignment of
  protein--protein interaction networks.
\newblock {\em Bioinformatics}, 29(13):1654--1662, 2013.

\bibitem{pachauri2013solving}
D.~Pachauri, R.~Kondor, and V.~Singh.
\newblock Solving the multi-way matching problem by permutation
  synchronization.
\newblock In {\em Advances in neural information processing systems}, pages
  1860--1868, 2013.

\bibitem{paresfluid}
F.~Par{\'e}s, D.~Garcia-Gasulla, A.~Vilalta, J.~Moreno, E.~Ayguad{\'e},
  J.~Labarta, U.~Cort{\'e}s, and T.~Suzumura.
\newblock Fluid communities: A competitive and highly scalable community
  detection algorithm.
\newblock {\em Complex Networks \& Their Applications VI}, pages 229--240,
  2018.

\bibitem{patro2012global}
R.~Patro and C.~Kingsford.
\newblock Global network alignment using multiscale spectral signatures.
\newblock {\em Bioinformatics}, 28(23):3105--3114, 2012.

\bibitem{peyre2019computational}
G.~Peyr{\'e}, M.~Cuturi, et~al.
\newblock Computational optimal transport.
\newblock {\em Foundations and Trends{\textregistered} in Machine Learning},
  11(5-6):355--607, 2019.

\bibitem{peyre2016gromov}
G.~Peyr{\'e}, M.~Cuturi, and J.~Solomon.
\newblock Gromov-wasserstein averaging of kernel and distance matrices.
\newblock In {\em International Conference on Machine Learning}, pages
  2664--2672, 2016.

\bibitem{raghavan2007near}
U.~N. Raghavan, R.~Albert, and S.~Kumara.
\newblock Near linear time algorithm to detect community structures in
  large-scale networks.
\newblock {\em Physical review E}, 76(3):036106, 2007.

\bibitem{sharan2006modeling}
R.~Sharan and T.~Ideker.
\newblock Modeling cellular machinery through biological network comparison.
\newblock {\em Nature biotechnology}, 24(4):427, 2006.

\bibitem{singh2008global}
R.~Singh, J.~Xu, and B.~Berger.
\newblock Global alignment of multiple protein interaction networks with
  application to functional orthology detection.
\newblock {\em Proceedings of the National Academy of Sciences}, 2008.

\bibitem{sinkhorn1967concerning}
R.~Sinkhorn and P.~Knopp.
\newblock Concerning nonnegative matrices and doubly stochastic matrices.
\newblock {\em Pacific Journal of Mathematics}, 21(2):343--348, 1967.

\bibitem{sturm2006geometry}
K.-T. Sturm et~al.
\newblock On the geometry of metric measure spaces.
\newblock {\em Acta mathematica}, 196(1):65--131, 2006.

\bibitem{vayer2018fused}
T.~Vayer, L.~Chapel, R.~Flamary, R.~Tavenard, and N.~Courty.
\newblock Fused {G}romov-{W}asserstein distance for structured objects:
  theoretical foundations and mathematical properties.
\newblock {\em arXiv preprint arXiv:1811.02834}, 2018.

\bibitem{vayer2018optimal}
T.~Vayer, L.~Chapel, R.~Flamary, R.~Tavenard, and N.~Courty.
\newblock Optimal transport for structured data.
\newblock {\em arXiv preprint arXiv:1805.09114}, 2018.

\bibitem{vijayan2015magna++}
V.~Vijayan, V.~Saraph, and T.~Milenkovi{\'c}.
\newblock {MAGNA}++: {M}aximizing accuracy in global network alignment via both
  node and edge conservation.
\newblock {\em Bioinformatics}, 31(14):2409--2411, 2015.

\bibitem{villani2008optimal}
C.~Villani.
\newblock {\em Optimal transport: {O}ld and new}, volume 338.
\newblock Springer Science \& Business Media, 2008.

\bibitem{wang2011detecting}
L.~Wang, T.~Lou, J.~Tang, and J.~E. Hopcroft.
\newblock Detecting community kernels in large social networks.
\newblock In {\em 2011 IEEE 11th International Conference on Data Mining},
  pages 784--793. IEEE, 2011.

\bibitem{xie2018fast}
Y.~Xie, X.~Wang, R.~Wang, and H.~Zha.
\newblock A fast proximal point method for {W}asserstein distance.
\newblock {\em arXiv preprint arXiv:1802.04307}, 2018.

\bibitem{xu2019gromov}
H.~Xu, D.~Luo, H.~Zha, and L.~Carin.
\newblock Gromov-wasserstein learning for graph matching and node embedding.
\newblock {\em arXiv preprint arXiv:1901.06003}, 2019.

\bibitem{yan2015consistency}
J.~Yan, J.~Wang, H.~Zha, X.~Yang, and S.~Chu.
\newblock Consistency-driven alternating optimization for multigraph matching:
  A unified approach.
\newblock {\em IEEE Transactions on Image Processing}, 24(3):994--1009, 2015.

\bibitem{yan2015matrix}
J.~Yan, H.~Xu, H.~Zha, X.~Yang, H.~Liu, and S.~Chu.
\newblock A matrix decomposition perspective to multiple graph matching.
\newblock In {\em ICCV}, pages 199--207, 2015.

\bibitem{yang2016comparative}
Z.~Yang, R.~Algesheimer, and C.~J. Tessone.
\newblock A comparative analysis of community detection algorithms on
  artificial networks.
\newblock {\em Scientific reports}, 6:30750, 2016.

\bibitem{yu2018generalizing}
T.~Yu, J.~Yan, Y.~Wang, W.~Liu, et~al.
\newblock Generalizing graph matching beyond quadratic assignment model.
\newblock In {\em NIPS}, pages 861--871, 2018.

\bibitem{zhang2015multiple}
J.~Zhang and S.~Y. Philip.
\newblock Multiple anonymized social networks alignment.
\newblock In {\em ICDM}, pages 599--608, 2015.

\end{thebibliography}

\newpage
\appendix
\section{Details of Algorithms}
\subsection{The GWL framework for different tasks}\label{apx1}
Based on Algorithms~\ref{alg:prox} and~\ref{alg:gwb}, our GWL framework achieve the graph partitioning and matching tasks in Figures~\ref{fig:gm}-\ref{fig:gps}. 
The schemes of GWL for these tasks are shown in Algorithms~\ref{alg:gm}-\ref{alg:gps}.
\begin{algorithm}[H]
\small{
	\caption{$\mathcal{S}=\text{GWL-GraphMatching}(G_s,G_t,\gamma)$}
	\label{alg:gm}
	\begin{algorithmic}[1]
	    \REQUIRE $G_s=G(\mathcal{V}_s,\bm{C}_s,\bm{\mu}_s)$, $G_t=G(\mathcal{V}_t,\bm{C}_t,\bm{\mu}_t)$, hyperparameter $\gamma$.
	    \STATE Initialize correspondence set $\mathcal{S}=\emptyset$.
	    \STATE $\widehat{\bm{T}}=\text{ProxGrad}(G_s,G_t,\gamma)$.
		\STATE \textbf{For} $v_i^s\in\mathcal{V}_s$ 
		\STATE \quad Find $j=\arg\max_j \widehat{T}_{ij}$, then $\mathcal{S}=\mathcal{S}\cup\{(v_i^s, v_j^t)\}$.
		\RETURN $\mathcal{S}$
	\end{algorithmic}
}
\end{algorithm}
\begin{algorithm}[H]
\small{
	\caption{$\{G_k\}_{k=1}^{K}=\text{GWL-GraphPartitioning}(G,\gamma,K)$}
	\label{alg:gp}
	\begin{algorithmic}[1]
	    \REQUIRE $G=G(\mathcal{V}, \bm{C}, \bm{\mu})$, hyperparameter $\gamma$, the number of clusters $K$.
	    \STATE Initialize a node distribution via (\ref{eq:wb}): $\bm{\mu}_{\text{dc}}=\text{interpolate}_{K}(\text{sort}(\bm{\mu}))$
	    \STATE Construct a disconnected graph $G_{\text{dc}}=G(\mathcal{V}_{\text{dc}},\text{diag}(\bm{\mu}_{\text{dc}}),\bm{\mu}_{\text{dc}})$, where $\mathcal{V}_{\text{dc}}=\{1,...,K\}$.
	    \STATE $\widehat{\bm{T}}=\text{ProxGrad}(G,G_{\text{dc}},\gamma)$.
		\STATE Initialize $\mathcal{V}_k=\emptyset$ for $k=1,...,K$.
		\STATE \textbf{For} $v_i\in\mathcal{V}$ 
		\STATE \quad Find $j=\arg\max_j \widehat{T}_{ij}$, then $\mathcal{V}_j=\mathcal{V}_j\cup\{v_i\}$.
		\STATE \textbf{For} $k=1,...,K$
		\STATE \quad Construct a adjacency matrix by selecting rows and columns: $\bm{C}_k=\bm{C}(\mathcal{V}_k,\mathcal{V}_k)$. 
		\STATE \quad Construct a node distribution by selecting elements and normalizing them: $\bm{\mu}_k=\frac{\bm{\mu}(\mathcal{V}_k)}{\|\bm{\mu}(\mathcal{V}_k)\|_1}$.
		\RETURN $\{G_k=G(\mathcal{V}_k,\bm{C}_k,\bm{\mu}_k)\}_{k=1}^{K}$
	\end{algorithmic}
}
\end{algorithm}
\begin{algorithm}[H]
\small{
	\caption{$\mathcal{S}=\text{GWL-MultiGraphMatching}(\mathcal{G},\gamma)$} 
	\label{alg:gms}
	\begin{algorithmic}[1]
		\REQUIRE A graph set $\mathcal{G}=\{G_m=G(\mathcal{V}_m, \bm{C}_m, \bm{\mu}_m)\}_{m=1}^{M}$, hyperparameter $\gamma$
		\STATE Initialize correspondence set $\mathcal{S}=\emptyset$, $K=\min\{|\mathcal{V}_m|\}_{m=1}^{M}$, $\bm{\omega}=[\frac{1}{M},..,\frac{1}{M}]$. 
		\STATE $\{\widehat{\bm{T}}_m\}_{m=1}^{M}=\text{GWB}(\{G_m\}_{m=1}^{M},\gamma,K,\bm{\omega})$.
		\STATE \textbf{For} $k=1,...,K$ 
		\STATE \quad$\bm{s}=\emptyset$
		\STATE\quad\textbf{For} $m=1,..,M$
		\STATE\quad\quad Find $i=\arg\max_i \widehat{T}^m_{ik}$, then $\bm{s}=\bm{s}\cup\{v_i^m\}$.
		\STATE\quad $\mathcal{S}=\mathcal{S}\cup\bm{s}$.
		\RETURN $\mathcal{S}$.
	\end{algorithmic}
}
\end{algorithm}
\begin{algorithm}[H]
\small{
	\caption{$\{\mathcal{G}_k\}_{k=1}^{K}=\text{GWL-MultiGraphPartitioning}(\mathcal{G},\gamma,K)$}
	\label{alg:gps}
	\begin{algorithmic}[1]
		\REQUIRE A graph set $\mathcal{G}=\{G_m=G(\mathcal{V}_m, \bm{C}_m, \bm{\mu}_m)\}_{m=1}^{M}$, hyperparameter $\gamma$, the number of clusters $K$.
		\STATE Initialize $\bm{\omega}=[\frac{1}{M},..,\frac{1}{M}]$.
		\STATE $\{\widehat{\bm{T}}_m\}_{m=1}^{M}=\text{GWB}(\{G_m\}_{m=1}^{M},\gamma,K,\bm{\omega})$.
		\STATE Initialize $\mathcal{V}_{k,m}=\emptyset$ for $k=1,..,K$ and $m=1,..,M$.
		\STATE \textbf{For} $m=1,..,M$
		\STATE \quad\textbf{For} $v_i^m\in\mathcal{V}_m$ 
		\STATE \quad\quad Find $j=\arg\max_j \widehat{T}^m_{ij}$, then $\mathcal{V}_{j,m}=\mathcal{V}_{j,m}\cup\{v_i^m\}$.
		\STATE \quad\textbf{For} $k=1,...,K$
		\STATE \quad\quad $\bm{C}_{k,m}=\bm{C}_m(\mathcal{V}_{k,m},\mathcal{V}_{k,m})$, and $\bm{\mu}_{k,m}=\frac{\bm{\mu}_m(\mathcal{V}_{k,m})}{\|\bm{\mu}(\mathcal{V}_{k,m})\|_1}$.
		\RETURN $\{\mathcal{G}_k\}_{k=1}^{K}$, where $\mathcal{G}_k=\{G_{k,m}=G(\mathcal{V}_{k,m},\bm{C}_{k,m},\bm{\mu}_{k,m})\}_{m=1}^{M}$.
	\end{algorithmic}
}
\end{algorithm}

\subsection{The scheme of S-GWL}\label{apx2}
Based on Algorithms~\ref{alg:gm}, \ref{alg:gms} and \ref{alg:gps}, we show the scheme of our S-GWL method for (multi-) graph matching in Algorithm~\ref{alg:sgwl}.
\begin{algorithm}[H]
\small{
	\caption{$\mathcal{S}=\text{S-GWL}(\mathcal{G}_0,\gamma,K,R)$}
	\label{alg:sgwl}
	\begin{algorithmic}[1]
		\REQUIRE A graph set with $M$ graphs, $i.e.$, $\mathcal{G}_0=\{G_m=G(\mathcal{V}_m, \bm{C}_m, \bm{\mu}_m)\}_{m=1}^{M}$, $\gamma$, the number of partitions $K$ and that of recursions $R$.
		\STATE Initialize correspondence set $\mathcal{S}=\emptyset$.
		\STATE Initialize the root collection of graph sets as $\mathsf{G}_0=\{\mathcal{G}_0\}$.
		\STATE \textbf{For} $r=1,...,R$ \hfill $\backslash\backslash$ \texttt{Recursive $K$-partition mechanism}
		\STATE \quad Initialize $\mathsf{G}_r=\emptyset$.
		\STATE \quad \textbf{For} each graph set $\mathcal{G}\in \mathsf{G}_{r-1}$
		\STATE \quad\quad $\{\mathcal{G}_k\}_{k=1}^{K}=\text{GWL-MultiGraphPartitioning}(\mathcal{G},\gamma,K)$.
		\STATE \quad\quad $\mathsf{G}_r=\mathsf{G}_r\cup \{\mathcal{G}_k\}_{k=1}^{K}$.
		\STATE \textbf{For} each graph set $\mathcal{G}\in\mathsf{G}_{R}$
		\STATE \quad \textbf{If} $M=2$ \hfill $\backslash\backslash$ \texttt{Two-graph matching}
		\STATE \quad\quad $\mathcal{S}_{tmp}=\text{GWL-GraphMatching}(G_s,G_t,\gamma)$, where $\mathcal{G}=\{G_s,G_t\}$.
		\STATE \quad \textbf{Else} \hfill $\backslash\backslash$ \texttt{Multi-graph matching}
		\STATE \quad\quad $\mathcal{S}_{tmp}=\text{GWL-MultiGraphMatching}(\mathcal{G},\gamma)$.
		\STATE \quad $\mathcal{S}=\mathcal{S}\cup\mathcal{S}_{tmp}$.
		\RETURN $\mathcal{S}$.
	\end{algorithmic}
}
\end{algorithm}

\subsection{Detailed complexity analysis for GWL and S-GWL}\label{apx3}
\textbf{Algorithms~\ref{alg:gm} and~\ref{alg:gms}} Suppose that we have a source graph with $V_s$ nodes and $E_s$ edges and a target graph with $V_t$ nodes and $E_t$ edges. 
The most time- and memory-consuming operation in Algorithm~\ref{alg:gm} is the $\bm{C}_s\bm{T}^{(n)}\bm{C}_t^{\top}$ in (\ref{eq:proximal}). 
Because $\bm{C}_s$ is with size $V_s\times V_s$ and $\bm{C}_t$ is with size $V_t\times V_t$, the computational time complexity of this step in the worst case is $\mathcal{O}(V_s^2V_t + V_sV_t^2)$ and its memory complexity is $\mathcal{O}(V_s^2 + V_t^2 + V_sV_t)$.
Taking advantage of the sparsity of edge, $\bm{C}_s\bm{T}^{(n)}\bm{C}_t^{\top}$ can be implemented by sparse matrix multiplications ($i.e.$, save $\bm{C}_s$, $\bm{C}_t$ as ``csr'' matrix in Python), whose computational time complexity and memory cost can be reduced to $\mathcal{O}(E_sV_t + V_sE_t)$ and $\mathcal{O}(V_sV_t)$\footnote{The memory complexity actually should be $\mathcal{O}(E_s + E_t+V_sV_t)$. Based on the sparsity of edge, we ignore the edge-related terms.}, respectively. 
Assuming that these two graphs are with comparable size, we ignore the number of graphs and the subscripts and rewrite the time and memory complexity as $\mathcal{O}(VE)$ and $\mathcal{O}(V^2)$, as shown in the ``GWL'' column of Table~\ref{tab:complex}. 

Algorithm~\ref{alg:gms} is a natural extension of Algorithm~\ref{alg:gm} based on GWB.
Suppose that we have $M$ graphs.
We assume that these graphs and the target barycenter graph are with comparable size. 
The computational time complexity of Algorithm~\ref{alg:gms} is $\mathcal{O}(MVE)$ and its memory complexity is $\mathcal{O}(MV^2)$.

\textbf{Algorithms~\ref{alg:gp} and~\ref{alg:gps}} The main difference between Algorithm~\ref{alg:gp} and Algorithm~\ref{alg:gm} is that the size of target graph is much smaller than that of source graph, $i.e.$, $K=V_t\ll V_s$ and $K=E_t$, because the target graph is disconnected, whose number of nodes indicates the number of partitions in the source graph.
According to the analysis above, the time and memory complexity of Algorithm~\ref{alg:gp} is $\mathcal{O}(E_sK + V_sK)$ and $\mathcal{O}(E_s+V_sK)$\footnote{Even if edges are sparse, $E_s$ is often comparable to $V_sK$. Therefore, different from the analysis for Algorithms~\ref{alg:gm} and~\ref{alg:gms}, here we do not ignore $E_s$.}. 
Ignoring the subscripts, we obtain the complexity shown in Table~\ref{tab:syn_gp}. 

Similarly, Algorithm~\ref{alg:gps} is an extension of Algorithm~\ref{alg:gp} for $M$ graphs, whose time and memory complexity is $\mathcal{O}(MK(E + V))$ and $\mathcal{O}(M(E+VK))$, respectively.

\textbf{Algorithm~\ref{alg:sgwl}}
Given $M$ graphs with comparable sizes, each of which has about $V$ nodes and $E$ edges, we can apply $R=\lfloor\log_K V\rfloor$ recursions.
In the $r$-th recursion, the $\mathsf{G}_r$ in Algorithm~\ref{alg:sgwl}) contains $K^r$ sub-graph sets.
If we assume that each partitioning operation partition a graph into $K$ sub-graphs with comparable sizes, the $m$-th sub-graph in each set should be with $\mathcal{O}(\frac{V}{K^r})$ nodes and $\mathcal{O}(\frac{E}{K^r})$ edges. 
For each sub-graph set, we calculate its barycenter graph by Algorithm~\ref{alg:gps}, thus, its time and memory complexity is $\mathcal{O}(MK(\frac{E}{K^r} + \frac{V}{K^r}))$ and $\mathcal{O}(\frac{M}{K^r}(E+VK))$, respectively. 
At the end of recursion, we obtain $K^R$ sub-graph sets. 
Each sub-graph is very small, with size $\mathcal{O}(\frac{V}{K^R})$. 
As long as $K^R$ is comparable to $V$, the computations in lines 8-13 of Algorithm~\ref{alg:sgwl} can be ignored compared with the computations in the recursions.

In summary, we run $\lfloor\log_K V\rfloor$ recursions, and in the $r$-th recursion we need to calculate $K^r$ barycenter graphs. 
The overall time complexity of S-GWL is $\mathcal{O}(MK(E+V)\log_K V)$, and its memory complexity is $\mathcal{O}(M(E+VK))$, respectively, as shown in Proposition~\ref{prop:complexity}.
Choosing $K=2$ and ignoring the number of graphs, we obtain the complexity shown in Table~\ref{tab:complex}.

\subsection{Usefulness of node prior}
\begin{figure}[t]
    \centering
    \subfigure[$\gamma=0.001$]{
    \includegraphics[width=0.3\linewidth]{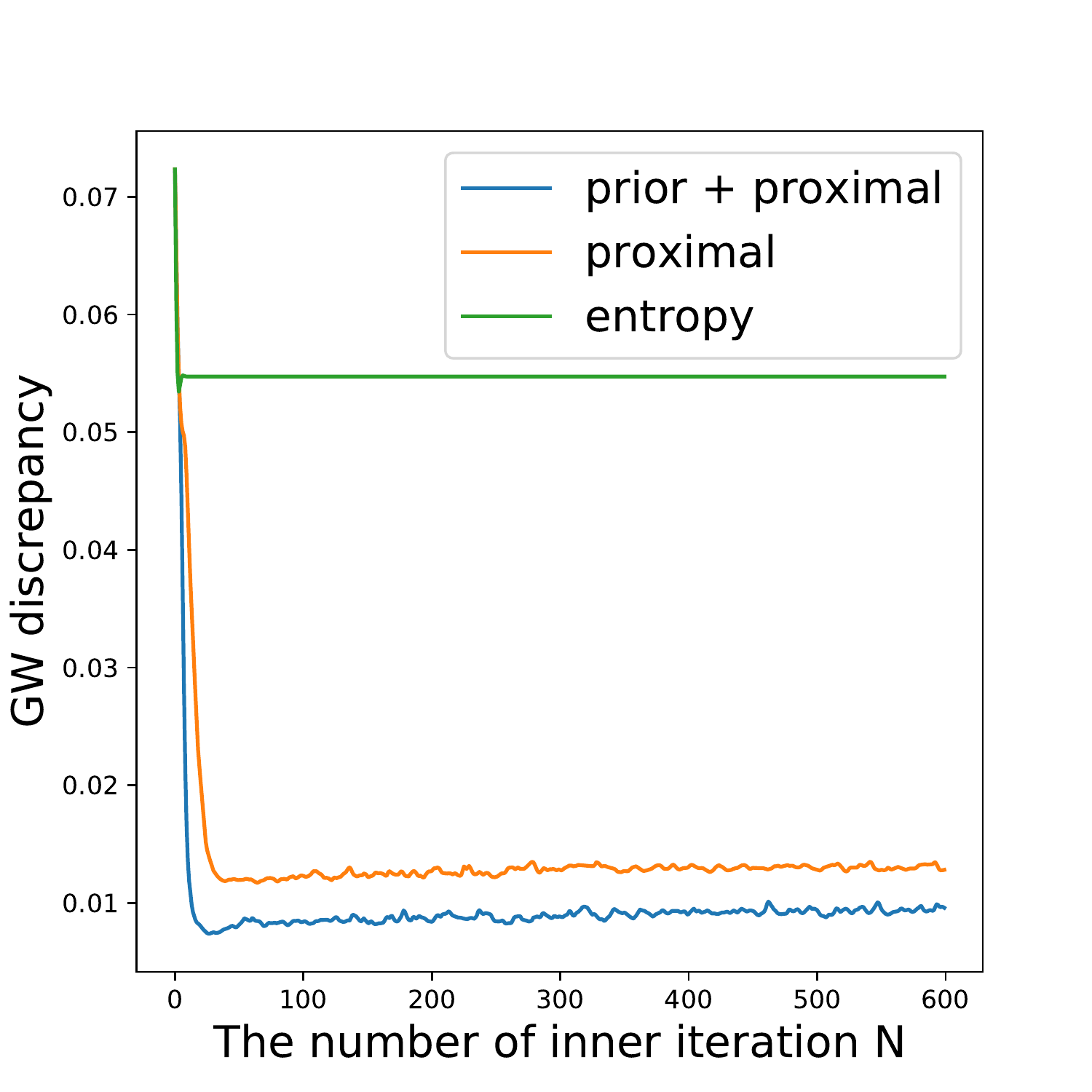}
    }\label{fig:c1}
    \subfigure[$\gamma=0.01$]{
    \includegraphics[width=0.3\linewidth]{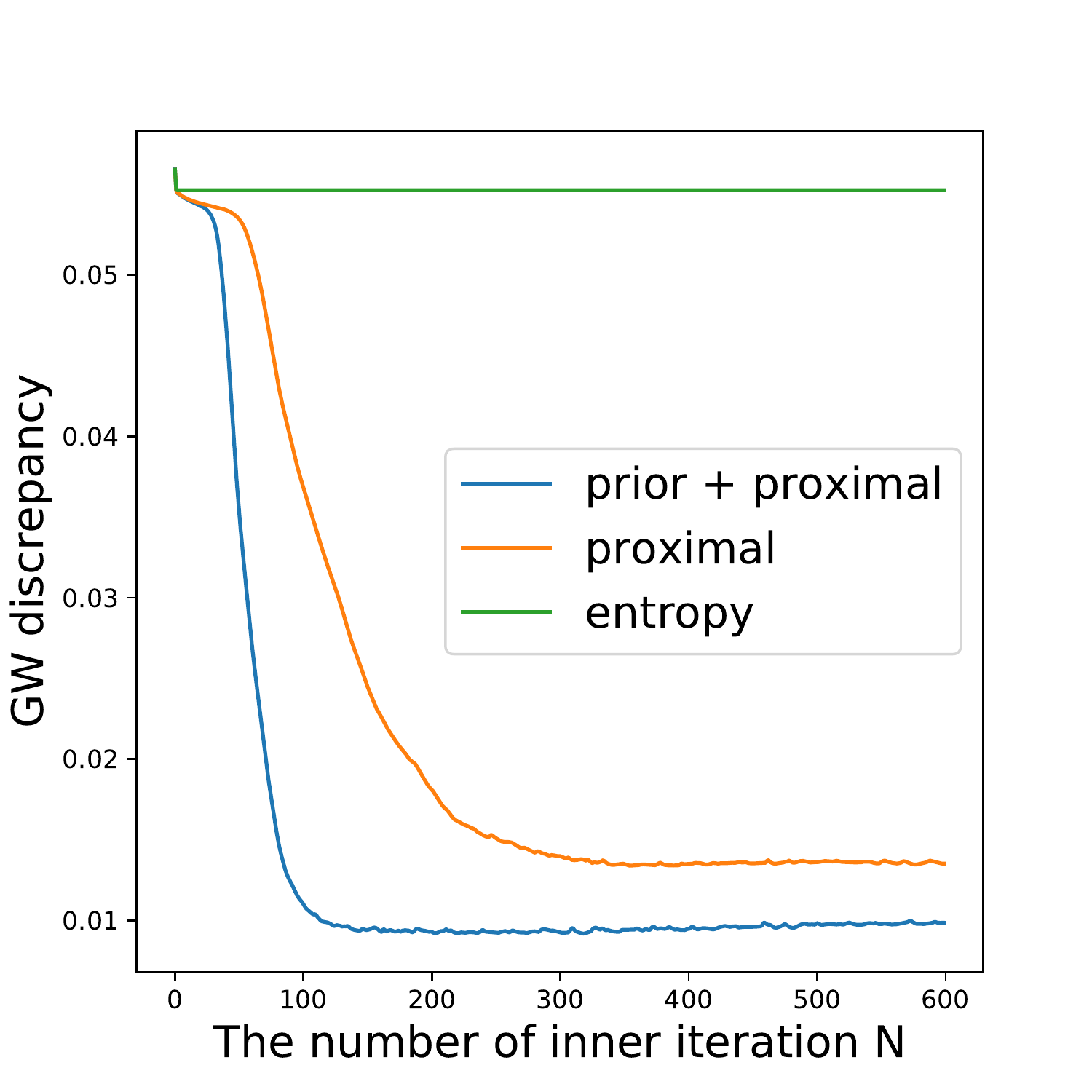}
    }\label{fig:c2}
    \subfigure[$\gamma=0.1$]{
    \includegraphics[width=0.3\linewidth]{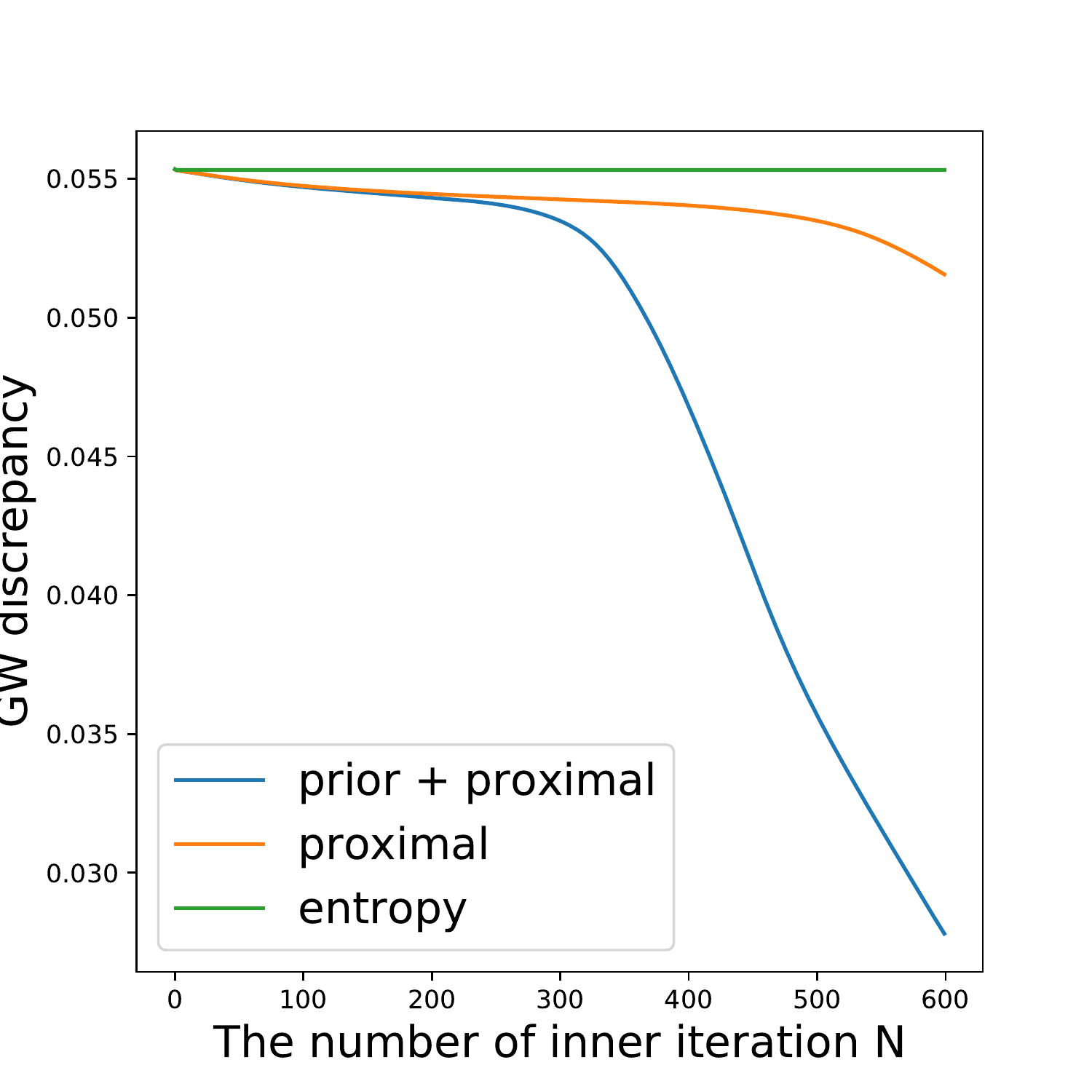}
    }\label{fig:c3}
    \vspace{-8pt}
    \caption{\small{Illustrations of the improvements on convergence achieved by our  proximal gradient method regularized by node prior ($i.e.$, ``prior + proximal'' compared with the entropy-based method in~\cite{peyre2016gromov}) and the vanilla proximal gradient method in~\cite{xu2019gromov}.}}\label{fig:node_prior}
\end{figure}

With the help of the prior knowledge of node ($i.e.$, $\bm{C}_{\text{node}}$), our regularized proximal gradient method can achieve a stable optimal transport with few iterations, whose rate of convergence is faster than the entropy-based method in~\cite{peyre2016gromov} and the vanilla proximal gradient method in~\cite{xu2019gromov}. 
Figure~\ref{fig:node_prior} illustrates the improvements on convergence achieved by our method. 
Given two synthetic graphs with 1,000 nodes, we calculate their GW discrepancy by different methods. 
Our method can reach lower GW discrepancy with fewer iterations, and its superiority is consistent with respect to the change of the hyperparameter $\gamma$.

\section{More Experimental Results}\label{apx4}
\subsection{Implementation details}
For each baseline, we list its source and language below:
\begin{itemize}
    \item Graph Partitioning:
    \begin{itemize}
        \item Metis (C): \url{http://glaros.dtc.umn.edu/gkhome/views/metis}
        \item FastGreedy (Python): \url{https://networkx.github.io/documentation/networkx-2.2/reference/algorithms/generated/networkx.algorithms.community.modularity_max.greedy_modularity_communities.html#networkx.algorithms.community.modularity_max.greedy_modularity_communities}
        \item Louvain (Python): \url{https://github.com/taynaud/python-louvain}
        \item Fluid (Python): \url{https://networkx.github.io/documentation/networkx-2.2/reference/algorithms/generated/networkx.algorithms.community.asyn_fluid.asyn_fluidc.html#networkx.algorithms.community.asyn_fluid.asyn_fluidc}
    \end{itemize}
    \item Graph Matching:
    \begin{itemize}
        \item PISwap (Python): \url{http://cb.csail.mit.edu/cb/piswap/webserver/}
        \item GHOST (C): \url{http://www.cs.cmu.edu/~ckingsf/software/ghost/} 
        \item MI-GRAAL (C): \url{http://www0.cs.ucl.ac.uk/staff/natasa/MI-GRAAL/index.html}
        \item MAGNA++ (C): \url{https://www3.nd.edu/~cone/MAGNA++/}
        \item HubAlign and NETAL (C):  \url{https://ttic.uchicago.edu/~hashemifar/}
        \item CPD+Emb (Python): node2vec is from \url{https://snap.stanford.edu/node2vec/}, CPD is from \url{https://github.com/siavashk/pycpd}.
        \item GWL+Emb (Python): \url{https://github.com/HongtengXu/gwl}.
    \end{itemize}
\end{itemize}
All the baselines are tested under their default settings. 
For our GWL framework and S-GWL method, their hyperparameters are set empirically in different experiments, which are shown in Table~\ref{tab:settings}.

\begin{table}[t]
\caption{The settings of hyperparameters in different experiments.}
\vspace{-6pt}
\centering
\small{
\setlength{\tabcolsep}{5pt}
\begin{tabular}{c|cccccc}
\hline\hline
Experiments &$\tau$ &$a$ &$b$ &$\gamma$ &$K$ &$R$\\ \hline
Synthetic partitioning (Table~\ref{tab:syn_gp}) &0 &0 &1 & 1e-2 &--- &---\\
EU-Email partitioning (Table~\ref{tab:real_gp}) &0 &0 &1e-3 & 5e-7 &--- &---\\
Indian-Village partitioning (Table~\ref{tab:real_gp}) &0 &5e-1 &1 & 5e-5 &--- &---\\
Synthetic matching (Figure~\ref{fig:2}) &1e1 &0 &1 & 2e-1 &2 &3\\
Yeast graph matching (Table~\ref{tab:syn_gm}) &1e3 &0 &1   &2.5e-2 &2 &3\\
MC3 network matching (Table~\ref{tab:syn_gm}) &1e1 &1 &1e-1 &1e-3 &2 &3\\
Yeast multi-graph matching (Table~\ref{tab:real_mgm}) &1e3 &0 &1 &2.5e-2 &8 &1\\
Yeast-Human matching (Table~\ref{tab:real_gm2}) &1 &0 &5e-1 &5e-2 &2 &4\\
\hline\hline
\end{tabular}\label{tab:settings}
}\vspace{-8pt}
\end{table}

\textcolor{black}{Note that using non-uniform node distributions is important for our method, especicially for the cases involving multi-graph partitioning and matching.
When doing multi-graph partitioning, the key step of our S-GWL, the adjacency matrix of the barycenter graph is initialized as a diagonal matrix and its node distribution is estimated by the node distributions of observed graphs. 
The node distribution based on node degree enhances the consistency of the partitioning across different graphs. 
For example, given two graphs $G_A$ and $G_B$, we jointly partition them into two subgraph pairs $\{G_A^1, G_B^1\}$ and $\{G_A^2, G_B^2\}$. 
If we use uniform node distributions, the barycenter will be initialized with uniform node distribution $[0.5, 0.5]^{\top}$ and adjacency matrix $0.5\bm{I}_2$, and we may have an identification problem --- $G_B^2$ can be finally paired with $G_A^1$.
}

\subsection{Performance on some challenging cases}
Although our GWL framework and S-GWL method perform well in most of our experiments, we find some challenging cases that point out our future research direction.

\begin{figure}[t]
    \centering
    \subfigure[Gaussian Partition: Accuracy v.s. efficiency]{
    \includegraphics[width=0.45\linewidth]{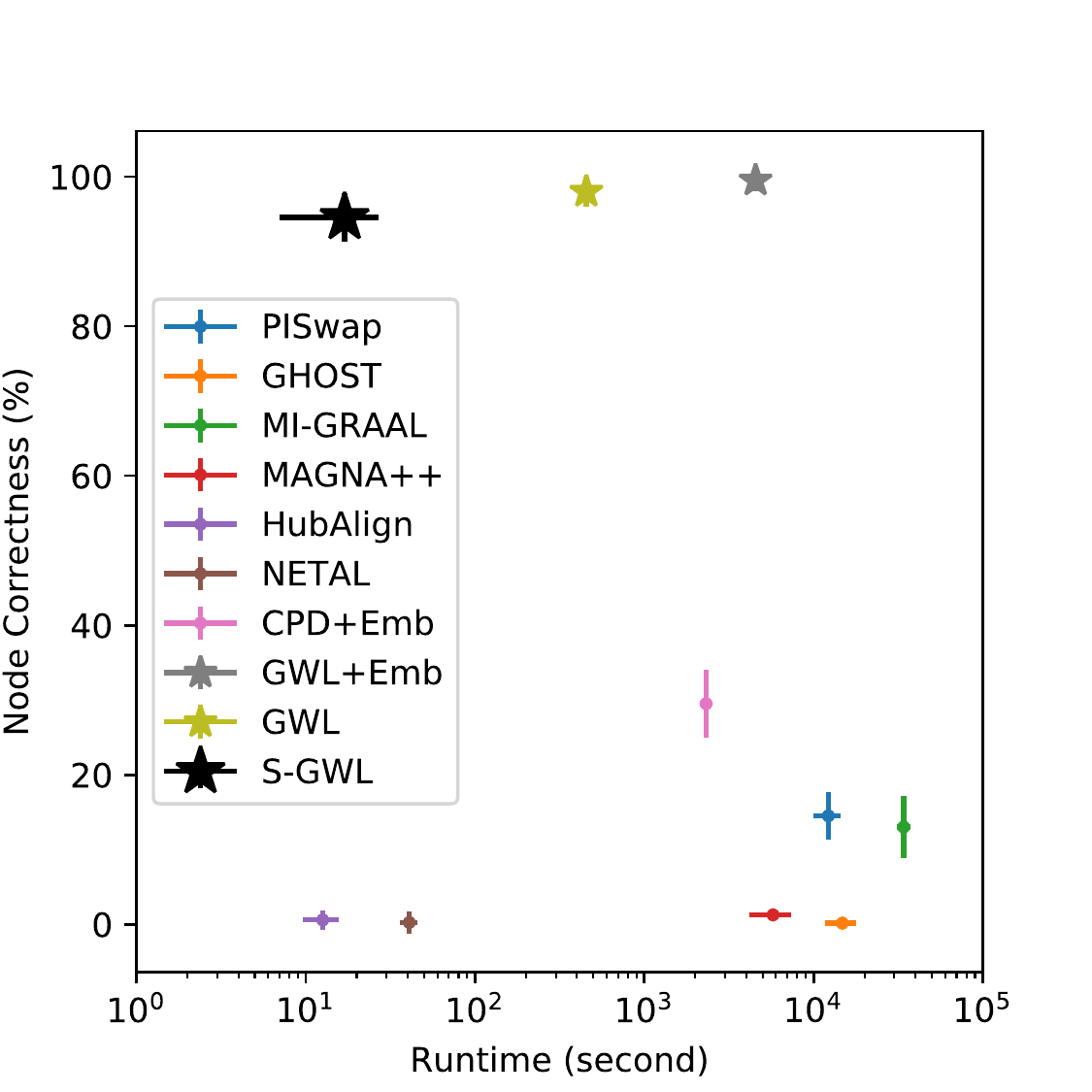}\label{fig:gauss1}
    }
    \subfigure[Barab{\'a}si-Albert: Accuracy v.s. efficiency]{
    \includegraphics[width=0.45\linewidth]{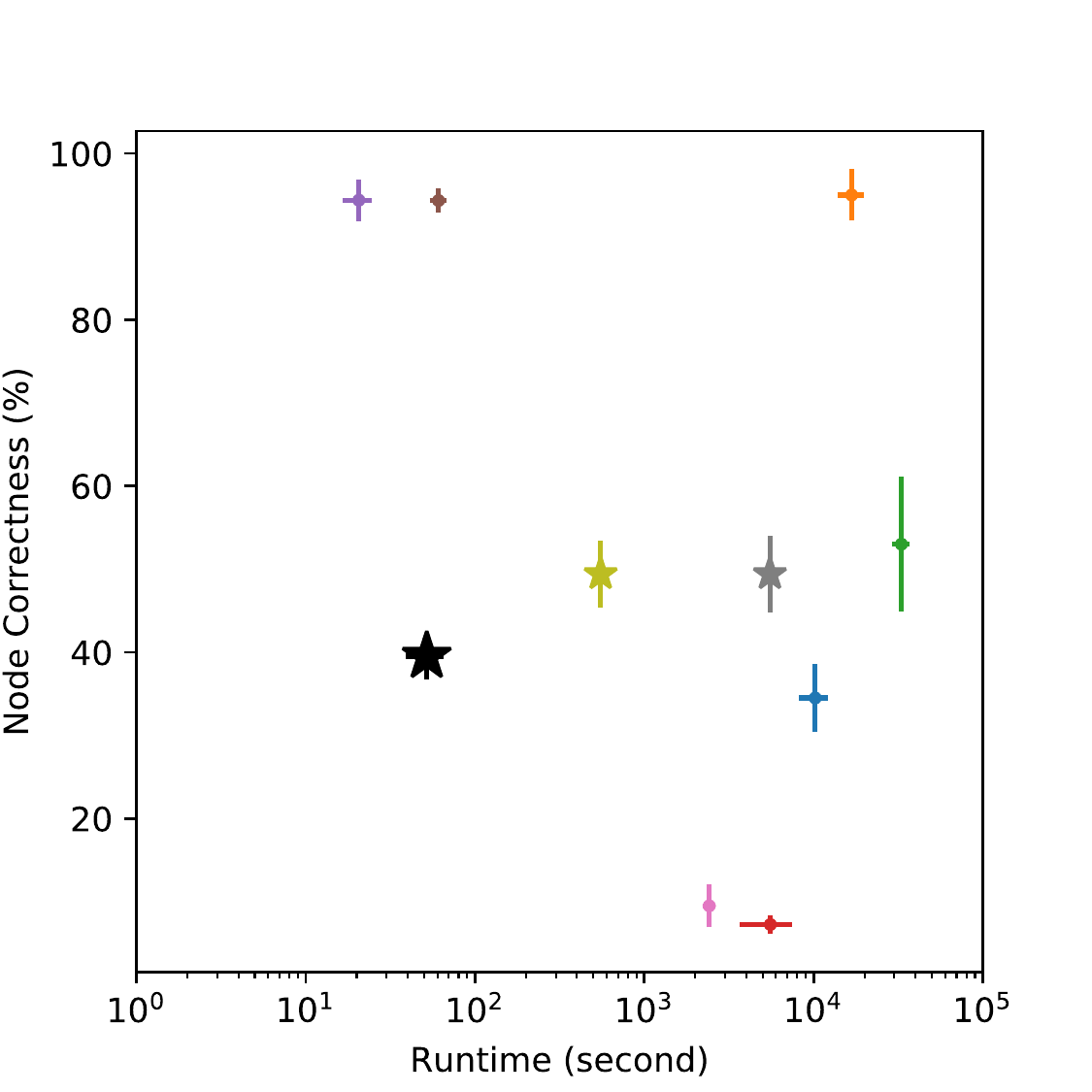}\label{fig:ba1}
    }
    \subfigure[Gaussian Partition: Acceleration]{
    \includegraphics[width=0.45\linewidth]{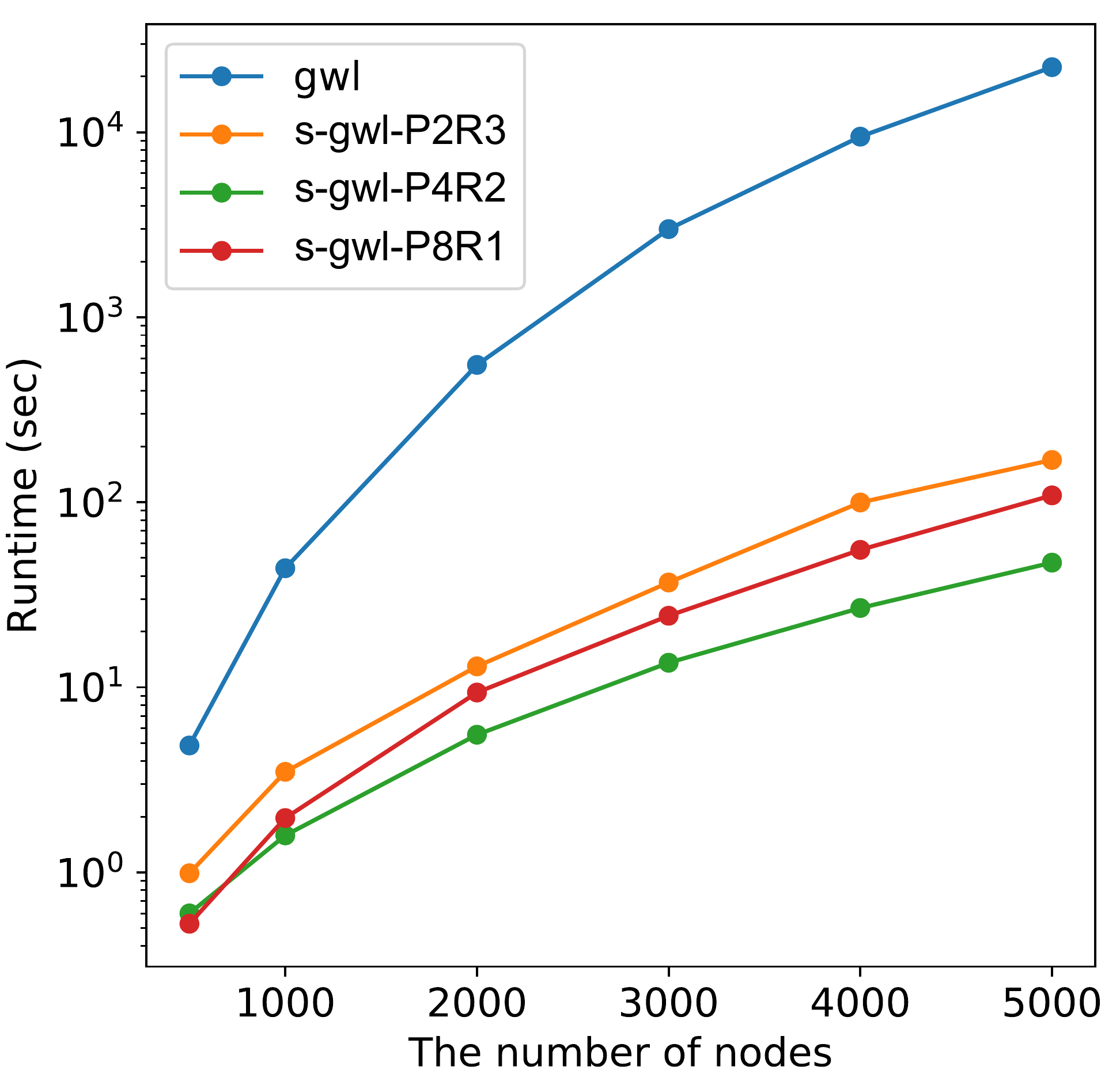}\label{fig:gauss2}
    }
    \subfigure[Barab{\'a}si-Albert: Acceleration]{
    \includegraphics[width=0.45\linewidth]{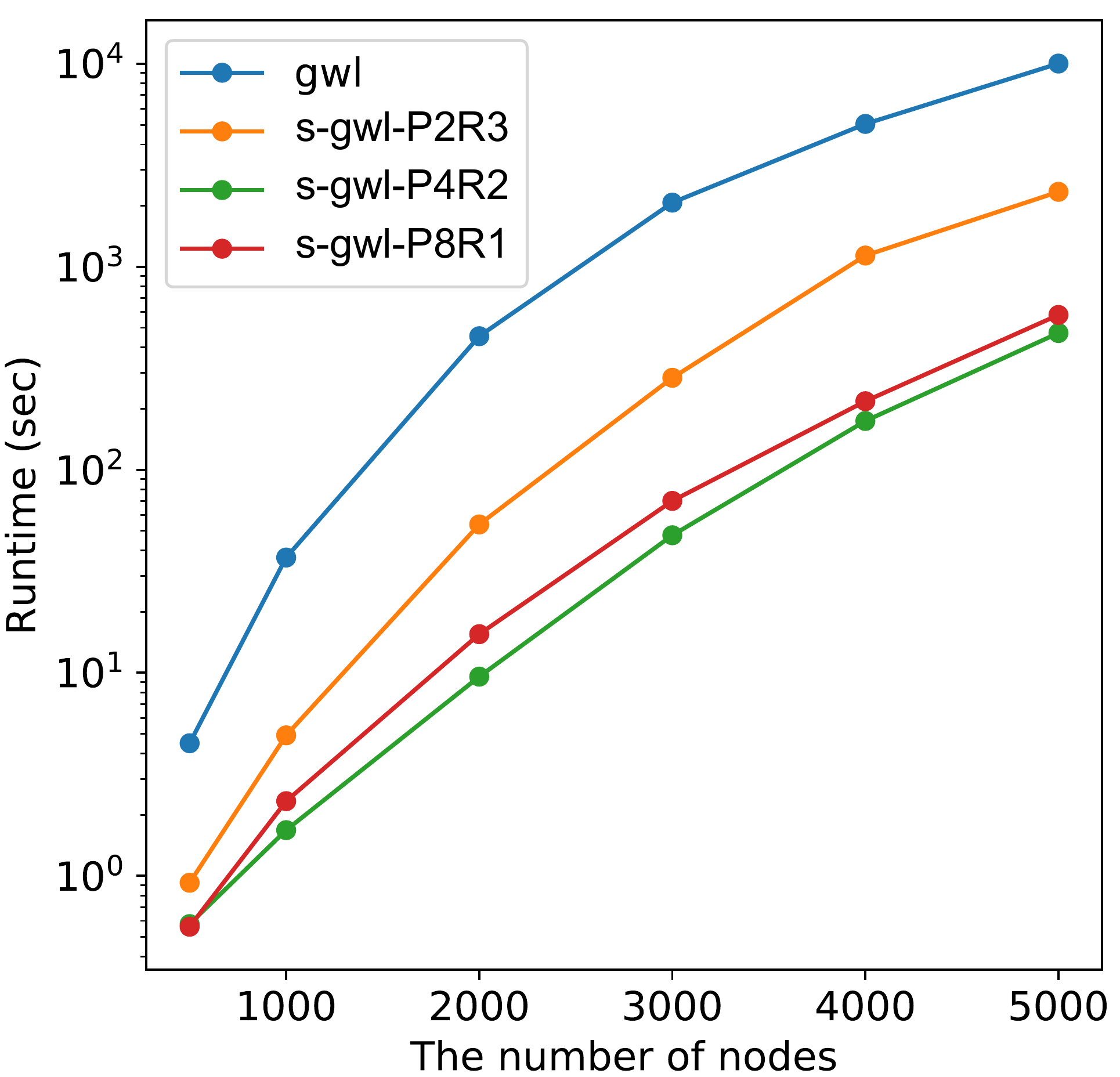}\label{fig:ba2}
    }
    \caption{\small{The performance of our method on different kinds of graphs. (a, b) For each method, its standard deviation of node correctness and that of runtime are shown as well.}
    }\label{fig:2}
\end{figure}

\textbf{Matching Barab{\'a}si-Albert (BA) graphs} 
Figure~\ref{fig:complexity} shows the averaged matching results in 10 trials.
In five of these trials, we match synthetic graphs obeying to Gaussian random partition model. 
In the remaining five trials, we match synthetic graphs obeying to Barab{\'a}si-Albert (BA) model. 
The overall performance shown in Figure~\ref{fig:complexity} demonstrates the superiority of our S-GWL method. 
This outstanding result is mainly contributed by the experiments on Gaussian partition graphs.
Specifically, when matching Gaussian partition graphs, all the GW discrepancy-based methods achieves very high node correctness, and the speed of our method is almost the same with the fastest HubAlign method, as shown in Figure~\ref{fig:gauss1}.
When it comes to BA graphs, Figure~\ref{fig:ba1} indicates that although GW discrepancy-based methods still outperform many baselines, there is a gap between them and the state-of-the-art methods in the aspect of node correctness. 

Additionally, the BA graphs also have a negative influence on our recursive mechanism. 
For Gaussian partition graphs, it is relatively easy to partition them into several sub-graphs with comparable size. 
In such a situation, the power of our recursive mechanism can be maximized, which helps us achieve over 100 times acceleration.
However, for BA graphs, the sub-graphs we get are often with incomparable size. 
The largest sub-graph decides the runtime of our S-GWL method. 
As a result, our S-GWL method only achieves about 10$\sim$20 times acceleration.

Currently, we are making efforts to improve the performance and the speed of our method on BA graphs. 
To solve this problem, we may need to use some node information, $e.g.$, introducing node embedding into our S-GWL method.

\textbf{Matching incomparable graphs} The second challenging case is matching incomparable graphs. 
This case is common in the field of bioinformatics, $e.g.$, matching the PPI networks from different species. 
When the networks are with incomparable size, the performance of GW discrepancy-based methods degrades.
For example, in Table~\ref{tab:real_gm2}, we match the PPI network of yeast to that of human. 
This yeast network has 2,340 proteins (nodes), while the human network has 9,141 proteins. 
Because the ground truth correspondence between these proteins is unknown, we use edge correctness to evaluate our method.
Specifically, edge correctness calculates the percentage of yeast's edges appearing in the human network.

Experimental results show that both GWL and S-GWL outperform most of their competitors except HubAlign and NETAL. 
The main reason for this phenomenon, in our opinion, is because the constraint of optimal transport. 
The constraint $\bm{T}\in \Pi(\bm{\mu}_s,\bm{\mu}_t)$ implies that each node in the target graph is assigned to a source node with a probability as long as its probability in $\bm{\mu}_t$ is nonzero. 
When the number of target nodes is much larger than that of source nodes, the real correspondence will be oversmoothed because each source node transports to too many target nodes.
To overcome this issue, we need to propose a preprocess to remove potentially-useless nodes from the large graph, which is another future work for us.

\begin{table}[!t]
\caption{Comparisons for graph matching methods on edge correctness (\%).}
\vspace{-6pt}
\centering
\begin{threeparttable}
\small{
\setlength{\tabcolsep}{5pt}
\begin{tabular}{c|cccccccc} 
\hline\hline
Method &IsoRank &PISwap &MI-GRAAL &GHOST &NETAL &HubAlign &GWL &S-GWL\\ \hline
Yeast$\leftrightarrow$Human &2.12 &2.16 &13.87 &17.04 &28.65 &21.59 &19.56 &18.89\\
\hline\hline
\end{tabular}\label{tab:real_gm2}
\vspace{-2pt}
\begin{tablenotes}
 \item\tiny{The results of baselines are from~\cite{hashemifar2014hubalign}.}
\end{tablenotes}
}\vspace{-8pt}
\end{threeparttable}
\end{table}

% \begin{table}[!t]
% \caption{Comparisons for multi-graph matching methods on yeast networks.}
% \vspace{-6pt}
% \centering
% \small{
% \setlength{\tabcolsep}{5pt}
% \begin{tabular}{c|cc|cc|cc|cc
% } 
% \hline\hline
% \multirow{2}{*}{Method} &
% \multicolumn{2}{c|}{3 graphs} &
% \multicolumn{2}{c|}{4 graphs} &
% \multicolumn{2}{c|}{5 graphs} &
% \multicolumn{2}{c}{6 graphs}\\ \cline{2-9}
% &
% NC@1 &NC@all &
% NC@1 &NC@all &
% NC@1 &NC@all &
% NC@1 &NC@all \\ \hline
% MultiAlign 
% &62.97 &45.19
% &--- &---
% &--- &---
% &--- &---\\
% GWL (Unknown $\bar{G}$)
% &46.63 &26.41
% &57.81 &20.66
% &60.05 &15.17
% &49.77 &7.39\\
% S-GWL (Unknown $\bar{G}$)
% &31.67 &27.87
% &40.84 &15.28
% &54.78 &13.49
% &61.25 &13.28\\
% GWL (Known $\bar{G}$)
% &\textbf{63.84} &\textbf{46.22}
% &\textbf{68.73} &\textbf{39.14}
% &71.61 &31.57
% &76.49 &28.39\\
% S-GWL (Known $\bar{G}$) 
% &60.06 &43.33
% &68.53 &38.45
% &\textbf{73.21} &\textbf{33.27}
% &\textbf{76.99} &\textbf{29.68}\\
% \hline\hline
% \end{tabular}\label{tab:real_mgm_full}
% }\vspace{-8pt}
% \end{table}

\end{document}